\documentclass[10pt]{article}

\usepackage{fullpage}
\usepackage{setspace}
\usepackage{parskip}
\usepackage{titlesec}
\usepackage[section]{placeins}
\usepackage{xcolor}
\usepackage{breakcites}
\usepackage{lineno}
\usepackage{hyphenat}
\usepackage{lscape} 
\usepackage{pifont}
\definecolor{darkolivegreen}{rgb}{0.32, 0.41, 0.2}
\newcommand{\cmark}{{\color{darkolivegreen} \ding{51}}}%
\newcommand{\xmark}{{\color{red} \ding{55}}}%

\PassOptionsToPackage{hyphens}{url}
\usepackage[colorlinks = true,
            linkcolor = blue,
            urlcolor  = blue,
            citecolor = blue,
            anchorcolor = blue]{hyperref}
\usepackage{etoolbox}

\usepackage[natbibapa]{apacite}

\renewenvironment{abstract}
  {{\bfseries\noindent{\abstractname}\par\nobreak}\footnotesize}
  {\bigskip}

\titlespacing{\section}{0pt}{*3}{*1}
\titlespacing{\subsection}{0pt}{*2}{*0.5}
\titlespacing{\subsubsection}{0pt}{*1.5}{0pt}

\usepackage{authblk}

\usepackage{graphicx}
\usepackage[space]{grffile}
\usepackage{latexsym}
\usepackage{textcomp}
\usepackage{longtable}
\usepackage{tabulary}
\usepackage{booktabs,array,multirow}
\usepackage{amsfonts,amsmath,amssymb}
\providecommand\citet{\cite}
\providecommand\citep{\cite}

\newif\iflatexml\latexmlfalse

\AtBeginDocument{\DeclareGraphicsExtensions{.pdf,.PDF,.eps,.EPS,.png,.PNG,.tif,.TIF,.jpg,.JPG,.jpeg,.JPEG}}

\usepackage[utf8]{inputenc}
\usepackage[english]{babel}

\usepackage{float}

\usepackage[margin=1.5in]{geometry}

\newcommand{\orcid}[1]{\href{https://orcid.org/#1}{\textcolor[HTML]{A6CE39}{#1}}}
\usepackage{subfig}

\usepackage[normalem]{ulem}

\usepackage{helvet}

\usepackage[bottom]{footmisc}

\newcommand{\mfrr}{\color{black}}

\begin{document}

\title{When and How to Fool Explainable Models (and Humans) with Adversarial Examples}

\author[*,1]{Jon Vadillo \orcid{0000-0002-7966-9561}}%
\author[1]{Roberto Santana \orcid{0000-0002-1005-8535}}%
\author[1,2]{Jose A. Lozano \orcid{0000-0002-4683-8111}}%
\affil[1]{University of the Basque Country UPV/EHU, 20018, Spain}%
\affil[2]{Basque Center for Applied Mathematics (BCAM), 48009, Spain}%
\affil[*]{Corresponding author.}%

\vspace{-1em}

  \date{}

\begingroup
\let\center\flushleft
\let\endcenter\endflushleft
\maketitle
\endgroup

\selectlanguage{english}
\begin{abstract}

Reliable deployment of machine learning models such as neural networks continues to be challenging due to several limitations. Some of the main shortcomings are the lack of interpretability and the lack of robustness against adversarial examples or out-of-distribution inputs. In this {\mfrr exploratory review}, we explore the possibilities and limits of adversarial attacks for explainable machine learning models. First, we extend the notion of adversarial examples to fit in explainable machine learning scenarios, in which the inputs, the output classifications and the explanations of the model's decisions are assessed by humans. Next, we propose a comprehensive framework to study whether (and how) adversarial examples can be generated for explainable models under human assessment, introducing and illustrating novel attack paradigms. In particular, our framework considers a wide range of relevant yet often ignored factors such as the type of problem, the user expertise or the objective of the explanations, in order to identify the attack strategies that should be adopted in each scenario to successfully deceive the model (and the human). The intention of these contributions is to serve as a basis for a more rigorous and realistic study of adversarial examples in the field of explainable machine learning.%
\end{abstract}%

\sloppy

\vspace{1.6cm}

\par\null

\section{Introduction}
\label{sec:introduction}

Machine learning models, such as deep neural networks, still face several weaknesses that hamper the development and deployment of these technologies, despite their outstanding and ever-increasing capacity to solve complex artificial intelligence problems. One of the main shortcoming is their black-box nature, which prevents analyzing and understanding their reasoning process, while such a requirement is ever more in demanded in order to guarantee a reliable and transparent use of artificial intelligence. To overcome this limitation, different strategies have been proposed in the literature \citep{zhang2021survey}, ranging from post-hoc explanation methods, which try to identify the parts, elements or concepts in the inputs that most affect the decisions of trained models \citep{zeiler2014visualizing, yosinski2015understanding, kim2018interpretability, ghorbani2019towards}, to more proactive approaches which pursue a transparent reasoning by training inherently interpretable models \citep{li2018deep, chen2019this, hase2019interpretable, alvarez-melis2018robust, saralajew2019classification, zhang2018interpretable}. 

Another issue that threatens the reliability of deep neural networks is their low robustness to adversarial examples \citep{szegedy2014intriguing,yuan2019adversarial}, 
that is, to inputs manipulated in order to maliciously change the output of a model while the changes are imperceptible to humans.
%
%
%which, 
Indeed, this
can be seen as a direct implication of their lack of human-like reasoning. Therefore, improving the explainability of the models is also a promising direction to achieve adversarial robustness, a hypothesis which is supported by recent works which show that interpretability and robustness are connected {\mfrr\citep{etmann2019connection,zhang2019interpreting,tsipras2019robustness,ros2018improving,noack2021empirical}}.

{\mfrr
Furthermore, the study of adversarial attacks against explainable models has gained interest in recent years, as will be fully reviewed in Sections \ref{sec:explanations_under_attack} and \ref{sec:connections_exp_adv}. In contrast to common adversarial attacks, which focus solely on changing the classification of the model \citep{yuan2019adversarial}, attacks on explainable models need to consider both changes in the classification and in the explanation supporting that classification. 
Another key difference when considering attacks against explainable models is related to their stealthiness. Generally, the only constraint assumed in order to produce a stealthy attack is that the changes added to the inputs must be imperceptible to humans.
However, the use of explainable models implies a different scenario, where it is assumed that a human will observe and analyze not only the input, but also the model classification and explanation. Therefore, uncontrolled changes in both factors may cause inconsistencies, alerting the human. For this reason, the assumption of explainable classification models introduces a new question regarding the definition of adversarial examples:} \emph{can adversarial examples be deployed if humans observe not only the input but also the output classification and/or the corresponding explanation?}

\subsection{Objectives and contributions}

\begin{figure}
    \centering
    \includegraphics[trim={0 1.9cm 9cm 0cm},clip,scale=0.87]{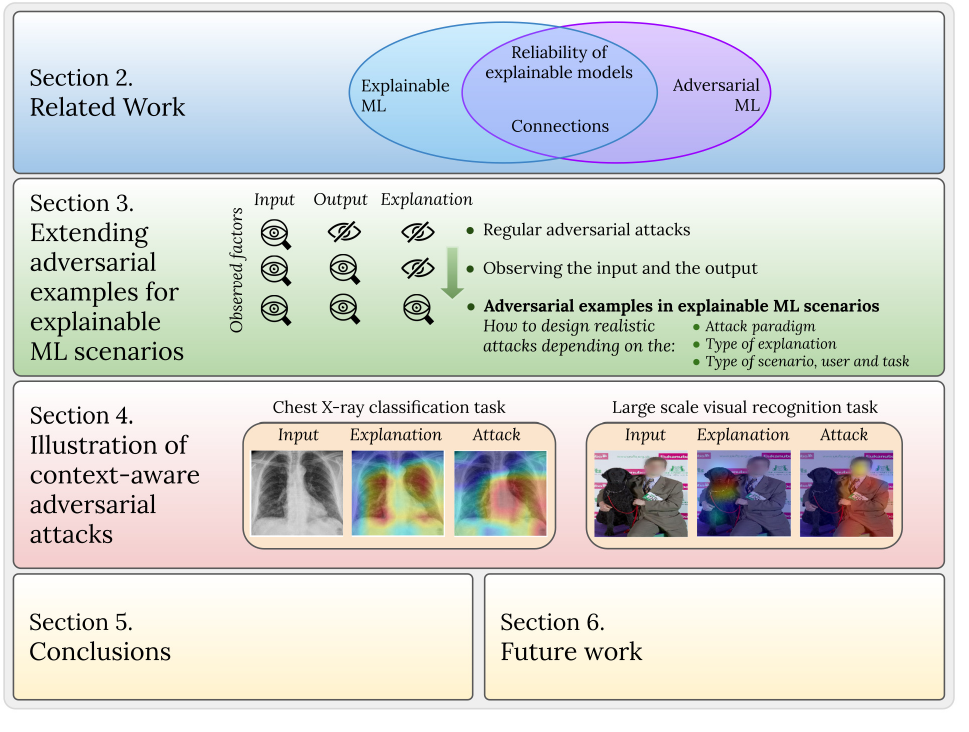}
    \caption{\mfrr Outline of our exploratory review. 
    }
    \label{fig:paper_outline}
\end{figure}

{\mfrr The objective of this exploratory review is to shed light}
on this question by extending the notion of adversarial examples for explainable machine learning scenarios, in which humans can not only assess the input sample, but also compare it to the output of the model and to the explanation. These extended notions of adversarial examples allow us to exhaustively analyze the possible attacks that can be produced by means of adversarially changing the model's classification and explanation, either jointly or independently (that is, changing the explanation without altering the output class, or vice versa). Our analysis leads to a comprehensive framework that establishes whether (and how) adversarial attacks can be generated for explainable models under human supervision. Moreover, we thoroughly describe the requirements that adversarial examples should satisfy in order to be able to mislead an explainable model (and even a human) depending on multiple scenarios or factors which, despite their relevance, are often overlooked in the literature of adversarial examples for explainable models, such as the expertise of the user or the objective of the explanation. Finally, the proposed attack paradigms are also illustrated by adversarial examples generated for two representative image classification tasks, as well as for two different explanation methods. {\mfrr The outline of our work is summarized in Figure \ref{fig:paper_outline}.}

%ae4xai_outline

The aim of all these contributions is to establish a basis for a more rigorous study of the vulnerabilities of explainable machine learning in adversarial scenarios. {\mfrr We believe that these fields will benefit from our work in the following ways. 

\begin{itemize}
    \item Heretofore, studies on adversarial attacks against explainable models have considered very particular or fragmented scenarios and attack paradigms. Thus, there is a lack of a unifying perspective in this field that connects all these works within a general analytical framework and taxonomy, which is a gap that we fill with this review.
    
    \item The framework we propose encompasses not only attack paradigms which have already been investigated in the literature, but also paradigms that, to the best of our knowledge, have not yet been studied, paving the way for new research venues.
    
    \item In addition, the role of the human is often overlooked in the study of attacks against explainable models, despite being a key factor in these scenarios. In this work, we address this limitation by thoroughly analyzing the requirements that adversarial examples should satisfy in order to be able to mislead an explainable model, and even a human, depending on the attack scenario. This analysis provides a comprehensive road map for the design of realistic attacks against explainable models.
    
    \item Furthermore, the fact that our framework considers a wide range of scenarios that an adversary may face allows us to summarize which paradigms are realistic or unrealistic in each of them, which is fundamental to ensure that attack methods are evaluated with an appropriate setting and methodology in future works.
    
    \item On another note, our work also contributes to raise awareness about the possible attack types that both models and humans may face in realistic adversarial scenarios, which is important to promote a more aware and secure use of machine learning based technologies, or even the development of more robust models or explanation methods. 
\end{itemize}

For the above reasons, the aim of this work is to contribute to a more methodical research in this area, delimiting the differences between the possible attack paradigms, identifying limitations in the current approaches and establishing more fine-grained and rigorous standards for the development and evaluation of new attacks or defenses.
}

\section{Related work}
\label{sec:related_work}

{\mfrr 
Our work focuses on adversarial attacks against explainable machine learning models. Therefore, this section provides a gentle yet comprehensive introduction to both research topics. This introduction will first summarize each research field independently, and, afterwards, the intersection between both, as described as follows.

To begin with, the fields of explainable and adversarial machine learning are presented in Sections \ref{sec:rw_explanation_methods} and \ref{sec:rw_adversarials}, respectively. Subsequently, the reliability of the explanation methods in adversarial scenarios is discussed in Section \ref{sec:explanations_under_attack}. Finally, further connections between explanation methods and adversarial examples are discussed in Section \ref{sec:connections_exp_adv}.
}

\subsection{Overview of explanation methods in machine learning}
\label{sec:rw_explanation_methods}

In  this section, we summarize the explanation methods proposed in the literature in order to present the terminology and taxonomy that will be used in the subsequent sections to develop our analytical framework on adversarial examples in explainable models.

\subsubsection{Scope, objective and impact of the explanations}

The objective of an explanation is to justify the behavior of a model in a way that is easily understandable to humans. However, different users might be interested in different aspects of the model, and, therefore, the explanations can be generated for different scopes or objectives. 

Overall, the scope of an explanation can be categorized as local or global \citep{zhang2021survey}. On the one hand, local methods aim to characterize or explain the model's prediction for each particular input individually, for example, by identifying the most relevant parts or features of the input.  On the other hand, global methods attempt to expose the general reasoning process of the model, for instance, summarizing (e.g., using a more simple but interpretable model) when a certain class will be predicted, or describing to what extent a particular input-feature is related to one class. Since in this paper we address the vulnerability of explainable models to adversarial examples, we focus on local methods.

In addition, explanations can be used, even for the same model, for different purposes. For instance, users querying the model for a credit loan might be interested in explaining the output obtained for their particular cases only, whereas a developer might be interested in discovering why that model misclassifies certain input samples. At the same time, an analyst can be interested in whether that model is biased against a social group for unethical reasons. At a higher level, all these purposes are based on necessities involving ethics, safety or knowledge acquisition, among others \citep{doshi-velez2018considerations}. Based on the purpose of the explanations and the particular problem, domain or scenario in which they are required, another relevant factor should be taken into consideration: the impact of the explanations, which can be defined as the consequence of the decisions made based on the analysis of the explanation. Healthcare domains are clear examples in which the consequences of the decisions can be severe. 

Despite the relevance of these factors, they are often overlooked when local explanation methods are designed or evaluated \citep{zhang2021survey,doshi-velez2018considerations}. The same happens for adversarial attacks in explainable models. We argue that the scope, the objective and the impact of explanations should be key factors when designing adversarial attacks against explainable models, since a different attack strategy needs to be adopted in each context to successfully deceive the model (and the human). This will be discussed in detail in Section \ref{sec:adv_for_explainable_ml}.

\subsubsection{Types of explanations}
\label{sec:explanation_types}

Different types of explanations exist depending on how the explanation is conveyed:

\begin{itemize}
\item \emph{Feature-based explanations:} assign an importance score to each feature in the input, based on their relevance for the output classification. Common feature-based explanations (especially in the image domain) are activation or saliency maps {\mfrr \citep{simonyan2014deep}}, which highlight the most relevant parts of the input. 
{\mfrr Despite their extensive use, } previous works have identified that such explanations can be unreliable and misleading \citep{kim2018interpretability,hase2019interpretable,chen2020concept,rudin2019stop,lipton2018mythos,kindermans2019reliability}. 

\item \emph{Example-based explanations:} the explanation is based on comparing the similarity between the input at hand and a set of \textit{prototypical} inputs that are representative of the predicted class. Thus, the classification of a given input sample is justified by the similarity between it and the prototypes of the predicted class. We will also refer to these types of explanations as \emph{prototype-based explanations} in the paper, although different forms of example-based explanation exist, such as the strategy proposed in \citet{koh2017understanding}, where influence functions are employed to estimate the training images \textit{most responsible} for a prediction.  Recent works have integrated prototype-based explanations directly in the learning process of neural networks, so that the classification is based on the similarities between the input and a set of prototypes \citep{li2018deep,chen2019this,hase2019interpretable,alvarez-melis2018robust}, achieving a more interpretable reasoning. The prototypes can represent an entire input describing one class (e.g., a prototypical handwritten digit "1" in digit classification) \citep{li2018deep}, or represent image-parts or semantic concepts \citep{chen2019this,alvarez-melis2018robust,hase2019interpretable}.

\begin{figure}[]
    \centering
    \subfloat[Feature-based explanation]{\includegraphics[scale=0.47]{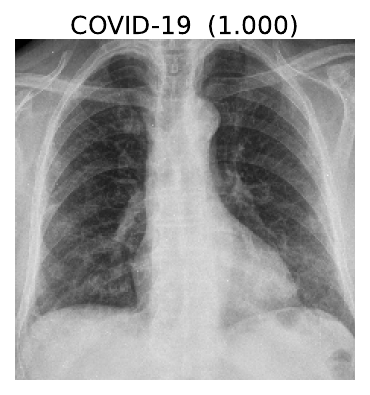}
    \hskip -8pt
    \includegraphics[scale=0.47]{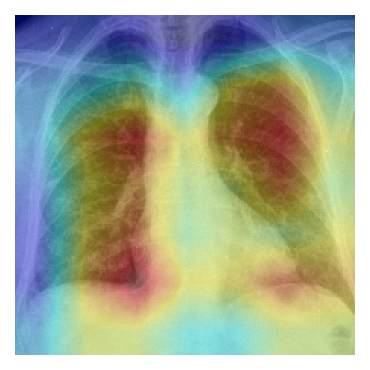}}
    \hskip 10pt
    \subfloat[Example-based explanation]{
    \includegraphics[scale=0.36]{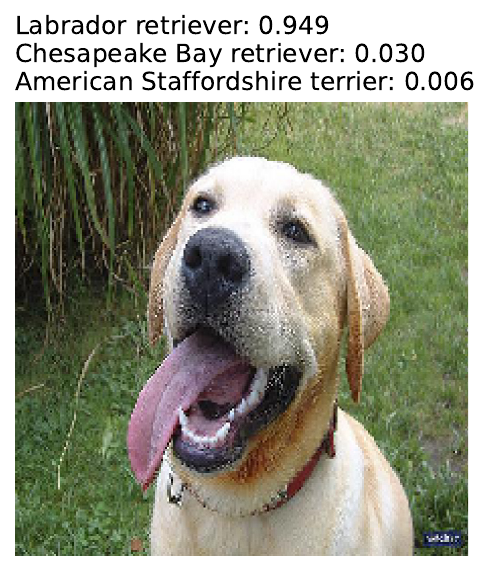} 
    \hskip -8pt
    \includegraphics[scale=0.37]{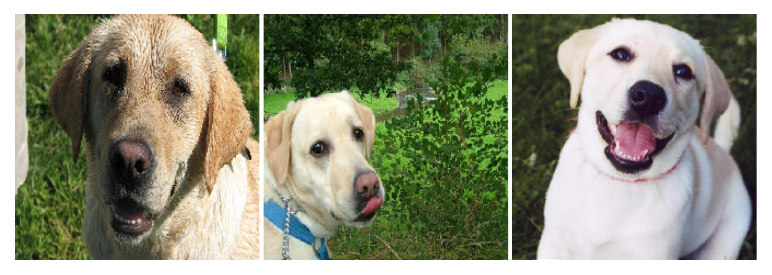}}\\
    \subfloat[Rule-based explanation]{\includegraphics[trim={0 7cm 14cm 0cm},clip,scale=0.46]{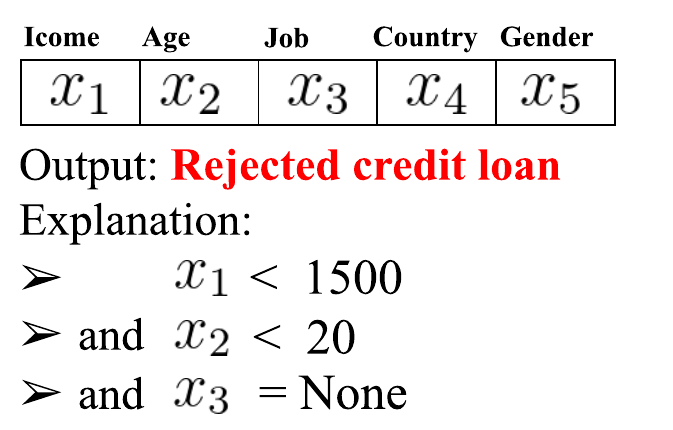}}
    \hskip 55pt
    \subfloat[Counterfactual explanation]{\includegraphics[trim={0 7cm 14cm 0cm},clip,scale=0.46]{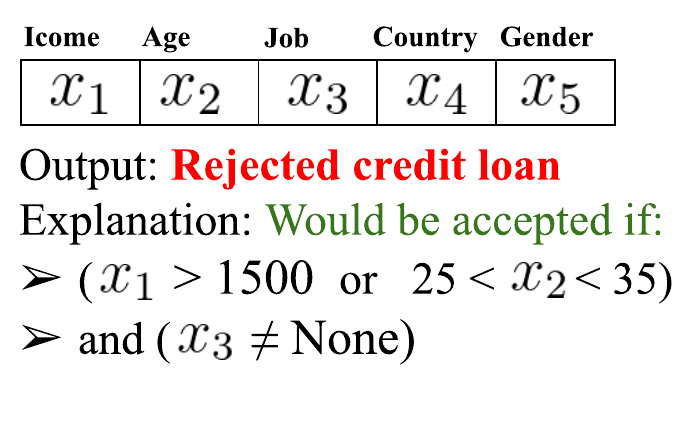}}
    \caption{\mfrr Illustrative examples of the four main types of explanations in machine learning.
    }
    \label{fig:illustrating_explanations}
\end{figure}

\item \emph{Rule-based explanations:} these explanation methods aim to expose the reasoning of a model in a simplified or human-understandable set of rules, such as logic-rules or if-then-else rules, which represent a natural form of explanations for humans {\mfrr \citep{lakkaraju2019faithful,vanderwaa2021evaluating}}. Rule-based explanations are particularly well-suited when the input contains features which are easily interpretable. 

\item \emph{Counterfactual explanations:} although counterfactual explanations \citep{wachter2017counterfactual} can be considered, in their form, as rule-based explanations, the main difference of these explanations is their conditional or hypothetical reasoning nature, as the aim is suggesting the possible changes that should happen in the input to receive a different (and frequently more positive) output classification (e.g., ``a rejected loan request would be accepted if the subject had a higher income'').

\end{itemize}

{\mfrr Some illustrative examples of these four types of explanations are presented in Figure \ref{fig:illustrating_explanations}.}
Overall, the most suitable type of explanation depends on the domain, the scope and the purpose of the explanation, as well as on the expertise level of the users querying the model. We refer the reader to \citet{zhang2021survey} and \citet{gilpin2018explaining} for a more fine-grained overview of explanation methods. 
{\mfrr These surveys also provide an exhaustive enumeration of relevant methods in the literature focused on computing such explanations.
}

{\mfrr
\subsection{Overview of adversarial attacks against machine learning models}
\label{sec:rw_adversarials}

Adversarial examples \citep{szegedy2014intriguing} are inputs deliberately manipulated by a malicious actor with the purpose of (i) fooling the model into providing incorrect predictions, and (ii) ensuring that the perturbations are imperceptible to humans. An illustration of an adversarial example is shown in Figure \ref{fig:illustrating_adversarials}.
The imperceptibility constraint ensures that the adversarial perturbation introduced to the inputs does not legitimate the change in the output classification. At the same time, the fact that imperceptible perturbations can fool machine learning models raised alarms regarding the vulnerability of machine learning models.
Indeed, adversarial examples have shown to be applicable to a wide range of high stakes and human-centered applications, such as healthcare systems  \citep{bortsova2021adversarial,finlayson2019adversarial,hirano2021universal,asgaritaghanaki2018vulnerability,joel2022using,ma2021understanding,paschali2018generalizability,yoo2020outcomes,li2020robust,paul2020mitigating,rahman2021adversarial,li2020anatomical},
surveillance systems \citep{wang2020transferable, thys2019fooling,xu2020adversarial,bai2021adversarial,sharif2016accessorize,zheng2020effective},
%For instance, adversarial examples against
machine translation \citep{belinkov2018synthetic,ebrahimi2018adversarial,zou2020reinforced,cheng2019robust,cheng2020seq2sick,zhao2018generating,michel2019evaluation},
dialogue or question and answering systems \citep{deng2021attacking,xue2020dpaeg},
social network based scenarios such as recommendation systems, spam detection or sentiment analysis  \citep{guo2021adversarial},
and financial applications such as credit loan approval or fraud detection systems \citep{fursov2021adversarial,cartella2021adversarial,ballet2019imperceptible,mathov2022not,sarkar2018robust,renard2019detecting,kumar2021evolutionary}.
The vulnerability to adversarial attacks has also been exposed in a wide range of popular \textit{machine learning as a service}
APIs \citep{borkar2021simple,papernot2017practical,ilyas2018blackbox}.

\begin{figure}[]
    \centering
    \subfloat[Original input]{\includegraphics[scale=0.5]{original_xray-eps-converted-to.pdf}}
    \subfloat[Adversarial example]{\includegraphics[scale=0.5]{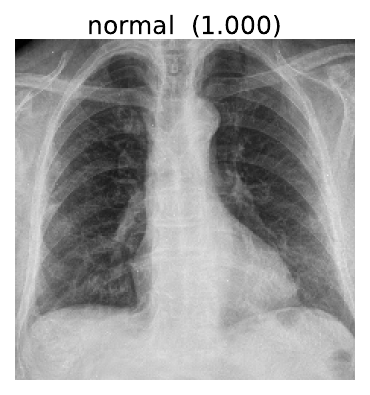}}
    \caption{\mfrr Illustration of an adversarial example generated for a chest X-ray (CXR) classification task, in which the objective is to categorize the status of the patient as one of three classes: ``normal'', Covid-19, or (non-Covid) pneumonia (more details in Section \ref{sec:tasks_datasets_models}). (a) Original input sample in which the patient is diagnosed with Covid-19. (b) Adversarially manipulated input, which is misclassified by the model as ``normal'' (i.e., no disease found in the patient) despite being perceptually identical to the original image.
    }
    \label{fig:illustrating_adversarials}
\end{figure}

\subsubsection{Taxonomy of adversarial attacks}

Different categories of attacks are considered in the literature depending on factors such as the specific type of error to be produced, the scope of the perturbation and the resources available to the adversary \citep{yuan2019adversarial}. In this section, we present the main categories in order to describe the most common attack paradigms studied in the literature. 

\begin{itemize}
\item \textbf{Type of misclassification}: First, two main types of attacks can be differentiated depending on whether the adversary aims to produce one particular incorrect class (targeted attacks), or simply a misclassification without any specific target class of preference (untargeted attacks).

\item \textbf{Scope of the perturbation}:
In addition, different types of perturbations can be considered depending on whether they are generated for one specific input at hand (individual perturbations) \citep{szegedy2014intriguing,goodfellow2015explaining,moosavi-dezfooli2016deepfool,carlini2017evaluating,madry2018deep} or whether they are designed to be input-agnostic and therefore effective with independence of the input in which they are applied (universal perturbations) \citep{moosavi-dezfooli2017universal,khrulkov2018art,mopuri2017fast,mopuri2018ask}. 

\item \textbf{Resources available to the adversary}: The information required by the adversary in order to effectively generate attacks leads to two main scenarios. On the one hand, in the white-box scenario, the adversary has full knowledge about the model internals (e.g., its architecture, weights or hyperparameters) and its training details. This allows highly efficient attacks to be generated, most of them relying on gradient-based strategies \citep{szegedy2014intriguing,goodfellow2015explaining,moosavi-dezfooli2016deepfool,carlini2017evaluating,madry2018deep,moosavi-dezfooli2017universal}. On the other hand, in black-box scenarios the adversary has no knowledge about the model \citep{ilyas2018blackbox,brendel2018decisionbased,alzantot2019genattack,papernot2017practical}. More intermediate scenarios (sometimes referred to as gray-box scenarios) can be assumed when the adversary has limited access to the models, such as the output confidences assigned to every possible class or the logit values \citep{ilyas2018blackbox}. The opacity in terms of model details requires more costly strategies than those used in the other case, such as evolutionary algorithms \citep{alzantot2019genattack,qiu2021blackbox}, gradient estimations \citep{chen2017zoo,ilyas2018blackbox}, or the use of surrogate models to generate the attack that are afterwards transferred to the initial model \citep{liu2017delving,papernot2017practical}.

\item \textbf{Type of deployment}:
Finally, a key aspect of adversarial examples is how those inputs are fed to the model. Generally, the assumed scenarios in research works allow the input to be modified ``digitally'', which is afterwards fed to the model. In other cases, \textit{physical} adversarial examples are crafted (e.g., printed traffic signs or malicious speech commands reproduced by a speaker) that are effective even when the signal is captured ``over-the-air'' and fed to the model \citep{xu2020adversarial,sharif2016accessorize,eykholt2018robust}. This allows circumventing the possible limitation in real-world scenarios in which the adversary might not have access to the digital files.

\end{itemize}

A summary of the taxonomy described can be consulted in Table \ref{tab:taxonomy_adv_attacks}. 
We also refer the reader to the work of \cite{yuan2019adversarial} for a more comprehensive and fine-grained survey on adversarial examples.

\begin{table}[]
\centering
\scalebox{0.82}{
{\mfrr
\begin{tabular}{|l|l|l|}
\hline
Factor &
  Categories &
  Explanation / goal \\ 
  \hline \hline
\multirow{2}{*}{\begin{tabular}[c]{@{}l@{}}
Type of\\ misclassification
\end{tabular}} &
  Targeted &
  Produce one particular incorrect output class. \\ \cline{2-3} 
 &
  Untargeted &
  \begin{tabular}[c]{@{}l@{}}
  Produce an incorrect class without any preference over the\\ available classes.\end{tabular} \\ 
  \hline \hline
\multirow{2}{*}{\begin{tabular}[c]{@{}l@{}}
Scope of the\\ Perturbation
\end{tabular}} &
  Individual &
  Optimized for one particular input. \\ 
  \cline{2-3} 
 &
  Universal &
  Optimized to be applicable to multiple inputs. \\ 
  \hline \hline
\multirow{2}{*}{\begin{tabular}[c]{@{}l@{}}
Resources \\ available to the\\ adversary\end{tabular}} &
  White-box scenario &
  \begin{tabular}[c]{@{}l@{}}
  The adversary has full knowledge of the model (e.g., weights,\\ hyperparameters, training details, etc.).
  \end{tabular} \\
  \cline{2-3} 
 &
  Black-box scenario &
  \begin{tabular}[c]{@{}l@{}}
  The adversary has none (or very limited) information about \\ the details of the model, which is considered a ``black-box''.
  \end{tabular} \\ 
  \hline \hline
\multirow{2}{*}{\begin{tabular}[c]{@{}l@{}}
Type of\\ deployment
\end{tabular}} &
  ``Digital-world'' &
  \begin{tabular}[c]{@{}l@{}}
  The adversarial example is crafted digitally (i.e., by\\ manipulating the ``digital file'') and is sent to the model ``as-is''.
  \end{tabular} \\ 
  \cline{2-3} 
 &
  ``Physical-world'' &
  \begin{tabular}[c]{@{}l@{}}
  The adversarial example is generated ``physically'' in order \\ to fool the model when the input is captured ``over-the-air''.
  \end{tabular} \\ 
  \hline
\end{tabular}
}
}
\caption{\mfrr Summary of the main taxonomy used to describe and categorize adversarial attacks.}
\label{tab:taxonomy_adv_attacks}
\end{table}

Finally, whereas the research on adversarial examples in the last few years has led to a torrent of attack methods proposed for multiple scenarios, tasks and even types of models, this research has focused almost exclusively on classification problems \citep{yuan2019adversarial}.
Nevertheless, it has been shown that adversarial examples can be generated for machine learning models trained to perform very different types of problems, such as regression \citep{balda2019perturbation,mode2020adversarial,tabacof2016adversarial,kos2018adversarial,gupta2021adversarial,li2020anatomical},
reinforcement learning \citep{hussenot2020copycat,lin2017tactics} or
image segmentation \citep{cisse2017houdini,fischer2017adversarial,metzen2017universal,mopuri2019generalizable,poursaeed2018generative,xie2017adversarial} problems. All these advances allow a wide range of opportunities for adversaries to maliciously take control of the outcomes of machine learning models, threatening countless systems.
At the same time, research has focused on models that only provide a classification as an answer. Only recently has the vulnerability of explainable models begun to be studied, as we will discuss in detail in the following section. 
}

\subsection{Reliability of explanations under adversarial attacks}
\label{sec:explanations_under_attack}

Some explanation methods in the literature have been proven to be unreliable in adversarial settings. In \citet{ghorbani2019interpretation}, \citet{dombrowski2019explanations}, \citet{alvarez-melis2018robustness}, \citet{zhang2020interpretable} and \citet{kuppa2020black}, it is shown that small changes in input samples can produce drastic changes in feature-importance explanations, while maintaining the output classification. In \citet{ghorbani2019interpretation}, the proposed attacks are also evaluated in the example-based explanations proposed in \citet{koh2017understanding}, based on estimating the relevance of each training image for a given prediction by using influence-functions. In \citet{zheng2019analyzing}, adversarial attacks capable of changing the explanations while maintaining the outputs are created for self-explainable (prototype-based) classifiers. In \citet{zhang2020interpretable} and \citet{kuppa2020black}, it is shown that adversarial examples can also produce wrong outputs and (feature-importance) explanations at the same time, or change the output while maintaining the explanations \citep{zhang2020interpretable}.

{\mfrr \citet{aivodji2019fairwashing} and \citet{lakkaraju2020how} show that} trustworthy explanations can be produced for a biased or an untrustworthy model, thus manipulating user trust. This approach is, however, not based on adversarial attacks, as they focus on producing a global explanation model that closely approximates the original (black-box) model but which employs trustworthy features instead of sensitive or discriminatory features (which are actually being used by the original model to predict). Similarly, in \citet{slack2020fooling}, \textit{adversarial} models are generated, capable of producing incorrect or misleading explanations without harming their predictive performance. In \citet{heo2019fooling}, a fine-tuning procedure is proposed to adversarially manipulate models, so that saliency map based explanations drastically change (becoming ineffective in highlighting the relevant regions), whereas the accuracy of the model is maintained.

Some works have also tried to justify the vulnerability of explanation methods to adversarial attacks, or the links between them. In \citet{ghorbani2019interpretation} and \citet{dombrowski2019explanations}, the non-smooth geometry of decision boundaries (of complex models) is blamed, arguing that, due to these properties, small changes in the inputs imply that the direction of the gradients (i.e., normal to the decision boundary) can drastically change. As most explanation methods rely on gradient information, the change in the gradient direction implies a different explanation. In \citet{zhang2020interpretable} and \citet{kuppa2020black}, the vulnerability is attributed to a gap between predictions and explanations. 
{\mfrr It is an open question whether this hypothesis holds for self-explainable models, which have been trained jointly to classify accurately and to provide explanations \citep{li2018deep, chen2019this, hase2019interpretable, alvarez-melis2018robust, saralajew2019classification}.}
Finally, theoretical connections between explanations and adversarial examples are established in \citet{etmann2019connection} and \cite{ignatiev2019relating}.
 
\subsection{Further connections between adversarial examples and interpretability} 
\label{sec:connections_exp_adv}

{\mfrr 
Paradoxically, using explanations to support or justify the prediction of a model can imply security breaches, as they might reveal sensitive information \citep{vigano2020explainable,sokol2019counterfactual}.}
For instance, an adversary can use explanations of how a black-box model works (e.g., what features are the most relevant in a prediction) in order to design more effective attacks. 
Similarly, in this paper we will show that justifying the classification of the model with an explanation makes it possible to generate types of deception using adversarial examples that, without explanations, it would not be possible to generate (e.g., to convince an expert that a misclassification of the model is correct).

On another note, recent works have shown that robust (e.g., adversarially trained) models are more interpretable \citep{etmann2019connection,zhang2019interpreting,tsipras2019robustness,ros2018improving}. In \citet{etmann2019connection}, this is justified by showing that the farther the inputs are with respect to the decision boundaries, the more aligned the inputs are with their saliency maps, thus, being more interpretable. {\mfrr Furthermore, \cite{noack2021empirical} show that enhancing the interpretability of a model during the training phase increases its adversarial robustness.}
{\mfrr Moreover, explanation methods have inspired particular defensive strategies against adversarial attacks \citep{renard2019detecting,jiang2021interpretabilityguided,liu2018adversarial,tao2018attacks,yang2020mlloo,zhang2018detecting,wang2020interpretability,hossam2021explain2attack}, and, inversely, adversarial attack methods have been proposed as a tool to generate or analyze explanations \citep{moore2019explaining,elliott2021explaining,haffar2021explaining,praher2021veracity}.
}

Finally, the similarities between interpretation methods and adversarial attacks and defenses are analyzed in \citet{liu2021adversarial}, showing how adversarial methods can be redefined from an interpretation perspective, and discussing how techniques from one field can bring advances into the other. Our paper, however, addresses a different objective. In contrast to \citet{liu2021adversarial}, which focuses on highlighting the similarities between particular methods from both fields, in this paper we propose a comprehensive framework to study if (and how) adversarial examples can be generated for explainable models under human assessment.

\section{Extending adversarial examples for explainable machine learning scenarios}
\label{sec:adv_for_explainable_ml}

In this section, we extend the notion of adversarial examples to fit in explainable machine learning contexts. For this purpose, {\mfrr in Section \ref{sec:adv_basic_case}}, we start from a basic definition of adversarial examples, and discuss more comprehensive scenarios in which the human subjects judge not only the input sample, but also the decisions of the model.
{\mfrr In Section \ref{sec:explanation_aware}, the human assessment of the explanations is also taken into account.}
To the best of our knowledge, no prior work has comprehensively addressed this type of generalization of adversarial examples.

This extended definition allows us to provide a general framework that identifies the way in which an adversary should design an adversarial example to deploy effective attacks even when a human is assessing the prediction process. The framework introduced also identifies the most effective ways of deploying attacks depending on factors such as {\mfrr the way in which the explanation is conveyed (Section \ref{sec:attacks_based_on_exp_types}) or the type of scenario, user and task (Section \ref{sec:scenarios}).} From an adversary perspective, this framework provides a comprehensive road map for the design of malicious attacks in realistic scenarios involving explainable models and a human assessment of the predictions. From the perspective of a developer or a defender, this road map helps to identify the most critical requirements that their explainable model should satisfy in order to be reliable.

\subsection{Scenarios in which human subjects are aware of the model predictions}
\label{sec:adv_basic_case}

\textit{Regular} adversarial examples are based on the assumption that an adversary can introduce a perturbation into an input sample, so that:
\begin{enumerate}
    \item The perturbation is not noticeable to humans, and, therefore, the human's perception of which class the input belongs to does not change.

    \item The class predicted by a machine learning model changes.
\end{enumerate}

Note that, according to this definition of adversarial examples, the human criterion is only considered regarding the input sample, without any human assessment of the model's output. {\mfrr However, this definition does not guarantee the stealthiness of the attack in scenarios in which the user observes the output classification, since the change in the output can be inconsistent, alerting the human.} 
For these reasons, the following question arises: \emph{are regular adversarial examples useful in practice when the user is aware of the output?}

\begin{figure}[!h]
    \centering
    \includegraphics[trim={0 8cm 8cm 0cm},clip,scale=0.7]{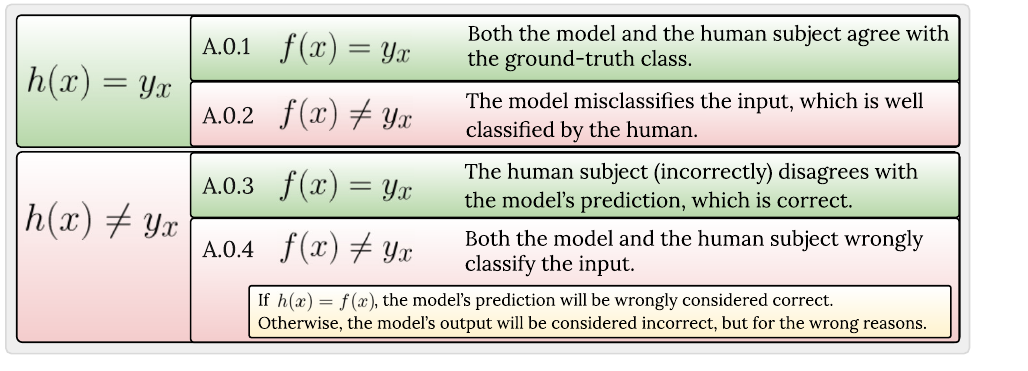}
    \caption{\mfrr Attack casuistry when the human observes not only the input but also the output classification of the model. 
    }
    \label{fig:output_casuistry}
\end{figure}

To address this question, we start by discussing four different scenarios, based on the agreement of the following factors:  $f(x)$, the model's prediction of the input; $h(x)$, the classification performed by a human subject; and $y_x$, the ground-truth class of an input $x$ (which will be unknown for both the model and the human subject in the prediction phase of the model). 
{\mfrr
For clarification, we assume that a human misclassification ($h(x)\neq y_x$) can occur in scenarios in which the addressed task is of high complexity, such as medical diagnosis \citep{pillai2019review}, or in which the label of an input is ambiguous, such as sentiment analysis \citep{agirre2006word,beck2020representation}.} 
{\mfrr Although a human misclassification might be uncommon in simple problems such as object recognition, even in such cases ambiguous or challenging inputs can be found  \citep{tsipras2020imagenet,stock2018convnets}}. Finally, unless specified, we will assume expert subjects, that is, subjects with knowledge in the task and capable of providing well-founded classifications.\footnote{Different degrees of expertise can be considered for a more comprehensive scenario, such as unskilled subjects, or partially skilled subjects capable of providing basic judgments about the input (for instance, a subject might not be able to visually discriminate between different species of reptiles, yet be able to visually classify an animal as a reptile and not as another animal class).} 
According to this framework, the four possible scenarios are {\mfrr those described in Figure \ref{fig:output_casuistry}.}

%output_casuistry

According to the described casuistry,  {\mfrr regular} adversarial attacks aim to produce the second scenario ({\mfrr A.0.2,} i.e., ${f(x)\neq h(x)=y_x}$), by imperceptibly perturbing an input $x_0$ that satisfies $f(x_0)=h(x_0)=y_{x_0}$ (i.e., the first scenario) so that the model's output is changed, but without altering the human perception of the input (which, therefore, implies $h(x)=y_x=y_{x_0}$).
However, assuming that the user is aware of the output, the fulfillment of the attack is subject to whether human subjects can correct the detected misclassification, or have control over the implications of that prediction. 
For example, an adversarial traffic sign will only produce a dramatic consequence in autonomous cars if the drivers do not take control with sufficient promptness.

Regarding the remaining cases, they do not fit in the definition of {\mfrr a regular} adversarial attack since either the input is misclassified by the human subject ($h(x)\neq y_x$) or the model is not fooled (${f(x)=h(x)=y_x}$). Nevertheless, assuming a more general definition, scenarios involving human misclassifications could be potentially interesting for an adversary. Similarly to regular adversarial attacks, which force the second scenario departing from the first one, an adversary might be interested in forcing the fourth scenario departing from the third one. 
{\mfrr
Let us take as an example a complex computer-aided diagnosis task through medical images, in which an expert subject fails in their diagnosis while the model is correct. In such cases, we can induce a human error confirmation attack by forcing the model to confirm the (wrong) medical diagnosis produced by the expert, that is, forcing $f(x)=h(x)\neq y_x$ \citep{johnson2019ai,goddard2012automation,bortsova2021adversarial,finlayson2019adversarial}.}

Based on the above discussion, we can determine that some types of adversarial attacks can still be effective even when the user is aware of the output. Nonetheless, paradoxically, it is possible to introduce new types of adversarial attacks when the output classification is supported by explanations, as we show in the following section.

\subsection{Scenarios in which human subjects are aware of the explanations}
\label{sec:explanation_aware}

The scenarios described in the previous section can be further extended for the case of explainable machine learning models, as the explanations for the predictions come into play. As a consequence, each of the cases defined above can be subdivided into new subcases depending on whether the explanations match the output class or whether humans agree with the explanations of the models. To avoid an exhaustive enumeration of all the possible scenarios, we focus only on those that we identify as interesting from an adversary perspective. From this standpoint, given an explainable model, adversarial examples can be generated by perturbing a well classified input (for which the corresponding explanation is also correct and coherent) with the aim of changing {\mfrr (i)} the output class, {\mfrr (ii)} the provided explanation or {\mfrr (iii)} both at the same time.

To formalize these scenarios, an explanation $A(x)$ will be defined as a set $A(x)=\{\phi_1, \phi_2, \dots, \phi_k \}$, where each $\phi_i$ represents a single explanatory unit which justifies or explains a decision in a human-understandable way. Let us denote $A_f(x)$ as the explanation provided to characterize the decision $f(x)$ of a machine learning model, and $A_h(x)$ as the explanation provided by a human according to their knowledge or criteria. 
The disagreement between $A_f(x)$ and $A_h(x)${\mfrr, denoted as $A_f(x)\not\approx A_h(x)$,} will be formalized as ${A_f(x) \cap A_h(x)=\emptyset}$, that is, as the lack of common explanatory units in both explanations. The total agreement between the explanations will be denoted as $A_f(x)=A_h(x)$. However, a total agreement is unlikely due to the high number of possible explanations for a given classification. In order to relax this definition, we will consider that there is an agreement between $A_f(x)$ and $A_h(x)$, which will be denoted as $A_f(x) \approx A_h(x)$, when both explanations overlap, that is, $A_f(x) \approx A_h(x) \Leftrightarrow A_f(x)\cap A_h(x) \neq \emptyset$. 
Similarly, we will denote $A(x) \sim y$ if an explanation $A(x)$ for the input $x$ is consistent with the reasons that characterize the class $y$ (that is, if the explanation correctly characterizes or supports the classification of $x$ as the class $y$). {\mfrr Formally, $A(x) \sim y \Leftrightarrow A(x)\subseteq A_Y(x) $, being $A_Y(x)$ the set of all explanatory units that justify classifying $x$ as $y$.}

For simplification, unless specified, we assume that given an input $x$ belonging to the class $y_x$, $h(x)=y_x$ and  $A_h(x) \sim  y_x$, this is, the human classification of an input into one class is correct and is based on reasons consistent for that class. Similarly, we will also assume that, for a \textit{clean} (unperturbed) input $x$, $f(x)=y_x$  and $A_f(x) \sim  y_x$.

The identified scenarios are as follows:

\begin{itemize}

    \item[A.1] $f(x) = y_x \ \wedge \ A_f(x)\not\approx A_h(x)$. In this case, the model is right but the explanations are incorrect or differ from those that would be provided by a human. Adversarial attacks capable of producing such scenarios have been studied in recent works for post-hoc feature-importance explanations \citep{ghorbani2019interpretation,dombrowski2019explanations,alvarez-melis2018robustness,zhang2020interpretable,kuppa2020black} and for self-explainable prototype-based classifiers \citep{zheng2019analyzing}, showing that small perturbations in the input can produce a drastic change in the explanations without changing the output. 
    \begin{itemize}
        \item[A.1.1] More particularly, we can imagine a scenario in which $A_f(x)\sim y_x$ despite ${A_f(x) \ \not\approx \ A_h(x)}$, for instance, if $A_f(x)$ points to relevant and coherent properties to classify the input as $y_x$, but which do not compose a correct or relevant explanation (with respect to the given input) according to a human criterion. From an adversary's perspective, changing the explanations without forcing a wrong classification allows confusing recommendations to be introduced. 
{\mfrr
For illustration, a model can correctly reject a loan request but the decision can be accompanied by a wrong {\mfrr yet coherent} explanation (e.g., ``the applicant is too young''), preventing the applicant from correcting the actually relevant deficiencies of the request (e.g., ``the applicant's salary is too low'') \citep{ustun2019actionable}.}
{\mfrr Similarly, a wrong explanation of a medical diagnosis system might lead to a wrong treatment or prescription \citep{ghassemi2021false,stiglic2020interpretability,bortsova2021adversarial,bussone2015role}.} 
In addition, biased or discriminative explanations could be produced with this attack scheme, for instance, attributing a loan rejection to sensitive features (e.g., gender, race or religion).
Such an explanation could make the models look unreliable or untrustworthy for users. 
{\mfrr 
Oppositely, biases could be hidden by producing trustworthy explanations to manipulate the trust of the users \citep{lakkaraju2020how,slack2020fooling,aivodji2019fairwashing,wang2020gradientbased}.} 
\end{itemize}

    \item[A.2] $f(x) \neq y_x \ \wedge \ A_f(x)\not\approx A_h(x)$. In this case, both the classification and the explanation provided by the model are incorrect. Adversarial attacks capable of producing such scenarios have been investigated in recent works \citep{zhang2020interpretable,kuppa2020black}. More particularly, we identify two specific sub-cases as relevant  when a human assesses the entire classification process:
    
    \begin{itemize}
        \item[A.2.1] $A_f(x) \sim f(x)$. In this case, the fact that the provided explanation is coherent with the (incorrectly) predicted class can increase the confidence of the human in the prediction, being therefore interesting from an adversary's perspective. We identify this case as the most direct extension of adversarial examples for explainable models, as the model is not only fooled but also supports its own misclassification with the explanation.  
        
        \item[A.2.2] $A_f(x) \not\sim f(x)$ {\mfrr $\wedge \ A_f(x) \not\sim y_x$}. This case is similar to the previous one (A.2.1), with the important difference that the model's explanation is now coherent with a class $y'$ different to $f(x)$ and $y_x$. 
        Thus, we are in a scenario in which a total mismatch is produced between all the considered factors. Whereas these attacks are an interesting case of study, they are also the most challenging to be deployed in practice without the inconsistencies being noticed. 
    \end{itemize}

    \item[A.3] $f(x)\neq y_x \ \wedge \ A_f(x)\approx A_h(x) \ \wedge \ A_f(x) \sim y_x$. In this case, the model's classification is wrong but the provided explanations are coherent from  a human perspective with respect to the ground-truth class $y_x$. The agreement in the explanations can increase the confidence in the model, but, at the same time, the output is not consistent with the explanation. However, the consistency issue might be solved by finding an input for which the explanation not only satisfies $A_f(x) \sim y_x$ but also $A_f(x) \sim f(x)$, for instance, by finding an ambiguous explanation that is applicable to both classes. {\mfrr Such attacks could be employed to convince the user to consider an incorrect class as correct or justified, or to bias the user's decision towards a preferred class (e.g., when there is more than one reasonable output class for an image).}

\end{itemize}

\begin{landscape}
\begin{table}[]
{\mfrr
\scalebox{0.68}{
\begin{tabular}{|l|c|ccc|ccc|l|l|}
\hline
\multirow{2}{*}{\begin{tabular}[c]{@{}l@{}}Factors\\ Observed\\ by the User\end{tabular}} &
  \multirow{2}{*}{ID} &
  \multicolumn{3}{c|}{Classification} &
  \multicolumn{3}{c|}{Explanation} &
  \multicolumn{1}{c|}{\multirow{2}{*}{\begin{tabular}[c]{@{}c@{}}Attack \\ Category / Description\end{tabular}}} &
  \multicolumn{1}{c|}{\multirow{2}{*}{\begin{tabular}[c]{@{}c@{}}Representative Examples,\\ Tasks or Use-cases\end{tabular}}} \\ \cline{3-8}
 &
   &
  \multicolumn{1}{c|}{\begin{tabular}[c]{@{}c@{}}Model\\ correct\end{tabular}} &
  \multicolumn{1}{c|}{\begin{tabular}[c]{@{}c@{}}Human\\ correct\end{tabular}} &
  \begin{tabular}[c]{@{}c@{}}Model-Human\\ agreement\end{tabular} &
  \multicolumn{1}{c|}{\begin{tabular}[c]{@{}c@{}}Model coherent\\ with ground-truth\end{tabular}} &
  \multicolumn{1}{c|}{\begin{tabular}[c]{@{}c@{}}Model coherent\\ with its output\end{tabular}} &
  \begin{tabular}[c]{@{}c@{}}Model-Human\\ agreement\end{tabular} &
  \multicolumn{1}{c|}{} &
  \multicolumn{1}{c|}{} \\ \hline
\multirow{2}{*}{\begin{tabular}[c]{@{}l@{}}Input+\\ Output\\(Sec. \ref{sec:adv_basic_case})\end{tabular}} &
  A.0.2 &
  \multicolumn{1}{c|}{\xmark} &
  \multicolumn{1}{c|}{\cmark} &
  \xmark &
  \multicolumn{1}{c|}{\textbf{--}} &
  \multicolumn{1}{c|}{\textbf{--}} &
  \textbf{--} &
  Regular attack. &
  \begin{tabular}[c]{@{}l@{}}Forcing misclassifications in critical\\ tasks (e.g., traffic-sign recognition, \\ surveillance or finance fraud detection).\end{tabular} \\ \cline{2-10} 
 &
  A.0.4 &
  \multicolumn{1}{c|}{\xmark} &
  \multicolumn{1}{c|}{\xmark} &
  \cmark &
  \multicolumn{1}{c|}{\textbf{--}} &
  \multicolumn{1}{c|}{\textbf{--}} &
  \textbf{--} &
  Human error confirmation. &
  \begin{tabular}[c]{@{}l@{}}Confirm a wrong diagnosis produced by \\ an expert in health-care domains.\end{tabular} \\ \hline
\multirow{6}{*}{\begin{tabular}[c]{@{}l@{}}Input+\\ Output+\\ Explanation\\(Sec. \ref{sec:explanation_aware})\end{tabular}} &
  A.1 &
  \multicolumn{1}{c|}{\multirow{2}{*}{\cmark}} &
  \multicolumn{1}{c|}{\multirow{2}{*}{\cmark}} &
  \multirow{2}{*}{\cmark} &
  \multicolumn{1}{c|}{\textbf{*}} &
  \multicolumn{1}{c|}{\textbf{*}} &
  \xmark &
  \begin{tabular}[c]{@{}l@{}}Incorrect explanation\\ (while keeping the correct\\ output).\end{tabular} &
  Reduce human trust in the model. \\ \cline{2-2} \cline{6-10} 
 &
  A.1.1 &
  \multicolumn{1}{c|}{} &
  \multicolumn{1}{c|}{} &
   &
  \multicolumn{1}{c|}{\cmark} &
  \multicolumn{1}{c|}{\cmark} &
  \xmark &
  \begin{tabular}[c]{@{}l@{}}Incorrect and coherent \\ explanations (while keep-\\ ing the correct output).\end{tabular} &
  \begin{tabular}[c]{@{}l@{}}Confusing recommendations in credit-loan \\ request or  medical-diagnosis tasks.\\ Biased or discriminative explanations.\\ Hide inappropriate behaviors of the model.\end{tabular} \\ \cline{2-10} 
 &
  A.2 &
  \multicolumn{1}{c|}{\multirow{3}{*}{\xmark}} &
  \multicolumn{1}{c|}{\multirow{3}{*}{\cmark}} &
  \multirow{3}{*}{\xmark} &
  \multicolumn{1}{c|}{\textbf{*}} &
  \multicolumn{1}{c|}{\textbf{*}} &
  \xmark &
  \begin{tabular}[c]{@{}l@{}}Incorrect output and \\ explanation.\end{tabular} &
  Reduce human trust in the model. \\ \cline{2-2} \cline{6-10} 
 &
  A.2.1 &
  \multicolumn{1}{c|}{} &
  \multicolumn{1}{c|}{} &
   &
  \multicolumn{1}{c|}{\xmark} &
  \multicolumn{1}{c|}{\cmark} &
  \xmark &
  \begin{tabular}[c]{@{}l@{}}Model is wrong but \\ supports its own \\ misclassification.\end{tabular} &
  \begin{tabular}[c]{@{}l@{}}Increase confidence of the human in the\\ incorrect prediction.\\ Bias the human in favour of a wrong\\ class.\end{tabular} \\ \cline{2-2} \cline{6-10} 
 &
  A.2.2 &
  \multicolumn{1}{c|}{} &
  \multicolumn{1}{c|}{} &
   &
  \multicolumn{1}{c|}{\xmark} &
  \multicolumn{1}{c|}{\xmark} &
  \xmark &
  \begin{tabular}[c]{@{}l@{}}Total mismatch between\\ the input, the classification\\ and the explanation.\end{tabular} &
  Reduce human trust in the model. \\ \cline{2-10} 
 &
  A.3 &
  \multicolumn{1}{c|}{\xmark} &
  \multicolumn{1}{c|}{\cmark} &
  \xmark &
  \multicolumn{1}{c|}{\cmark} &
  \multicolumn{1}{c|}{\cmark} &
  \cmark &
  \begin{tabular}[c]{@{}l@{}}Incorrect output while\\ keeping a correct\\ explanation.\end{tabular} &
  \begin{tabular}[c]{@{}l@{}}Ambiguous explanations applicable to\\ more than one class. \\ Misdirect the attention of the user towards\\ another reasonable class.\end{tabular} \\ \hline
\end{tabular}
}
}
\caption{\mfrr{ Overview of the attack casuistry described in Sections \ref{sec:adv_basic_case} and \ref{sec:explanation_aware}.
For the sake of simplicity, we use the following symbols to represent the following terms: \cmark \ (yes), \ \xmark \ (no), \textbf{--} (not applicable). {\mfrr In those paradigms in which subcases are considered, the symbol \textbf{*} is used to represent the term ``not specified'' (i.e., the choice made for those factors determines the attack subtype).}
}}
\label{tab:attack_casuistry}
\end{table}
\end{landscape}

\subsubsection{Attack design based on the type of explanation}
\label{sec:attacks_based_on_exp_types}
{\mfrr
Whereas our framework considers the explanations of the models in their most general form, the way in which an explanation is conveyed determines how humans process and interpret the information \citep{doshi-velez2018considerations,zhang2021survey}.
}
This implies that some attack strategies might be more suitable for some types of explanations than for others. Moreover, the way in which an adversarial example is generated for an explainable machine learning model will also depend on the type of explanation. For these reasons, in this section we briefly discuss in which way an adversarial example should be designed depending on the type of explanation or the particular type of attack to be produced.

\begin{itemize}
     \item \emph{Feature-based explanations:} the highlighted parts or features need to be coherent with the classification, and correspond to (I) human-perceivable, (II) semantically meaningful and (III) relevant parts. A common criticism to feature-based explanations such as saliency maps is that they identify the relevant parts of the inputs, but not how the models are processing such parts \citep{rudin2019stop}. Thus, an adversarial attack could take advantage of this limitation.  First, a particular region of the input can be highlighted to support a misclassification of the model and to convince the user (assuming that the region contributes to predict an incorrect class), which is interesting particularly for targeted adversarial attacks. 
     An attack could also highlight irrelevant parts to mislead the observer, or generate ambiguous explanations, by highlighting multiple regions or providing a uniform map, which are strategies well-suited for untargeted attacks. 

    \item \emph{Prototype-based explanations:} in this case, for the human to accept the given explanation, the key features of the closest prototypes should (I) be perceptually identifiable in the given input, and, ideally, (II) contain features correlated with the output class. The contrary should happen for the farthest prototypes, that is, their key features should not be present in the input nor be correlated with the output class (or, ideally, be opposite).
    In order to achieve these objectives, the more general the prototypes (e.g., if they represent semantic concepts or parts of inputs rather than completely describing an output class), the higher the chances of producing explanations that could lead to a wrong classification while being coherent with a human perception, such as ambiguous explanations.

    \item \emph{Rule-based explanations}
    can be fooled by targeting explanations which are aligned with the output of the model (e.g., the explanation justifies the prediction or at least mimics the behavior of the model), but which employ reliable, trustworthy or neutral features {\mfrr \citep{lakkaraju2020how,aivodji2019fairwashing}. For instance, a model for criminal-recidivism prediction could provide a negative assessment based on unethical reasons, whereas the explanation is taken as ethical \citep{lakkaraju2020how,aivodji2019fairwashing}.}
     
    \item \emph{Counterfactual explanations:} in this case, the objective of an adversarial attack could be forcing a particular counterfactual explanation, suggesting changes on irrelevant features (thus preventing correcting the deficiencies which are actually relevant), or forcing biased or discriminatory explanations in detriment to the fairness of the model.
     
\end{itemize}

\subsubsection{Attack design based on the scenario, user and task}
\label{sec:scenarios}

{\mfrr To conclude our framework,} we describe the main characteristics or desiderata that an adversarial attack should satisfy in different scenarios in order to be successful.
We build on the idea that common tasks, problems or applications share common categories, and that explanations or interpretation needs are different in each of them \citep{doshi-velez2018considerations}. Thus, adversarial attacks (or, oppositely, the defensive countermeasures) should also be designed differently for each type of explanation, focusing on the most relevant or crucial factor in each case.

\begin{figure}[!b]
    \centering
    \includegraphics[trim={3cm 4.1cm 2.5cm 2.7cm},clip,scale=0.5]{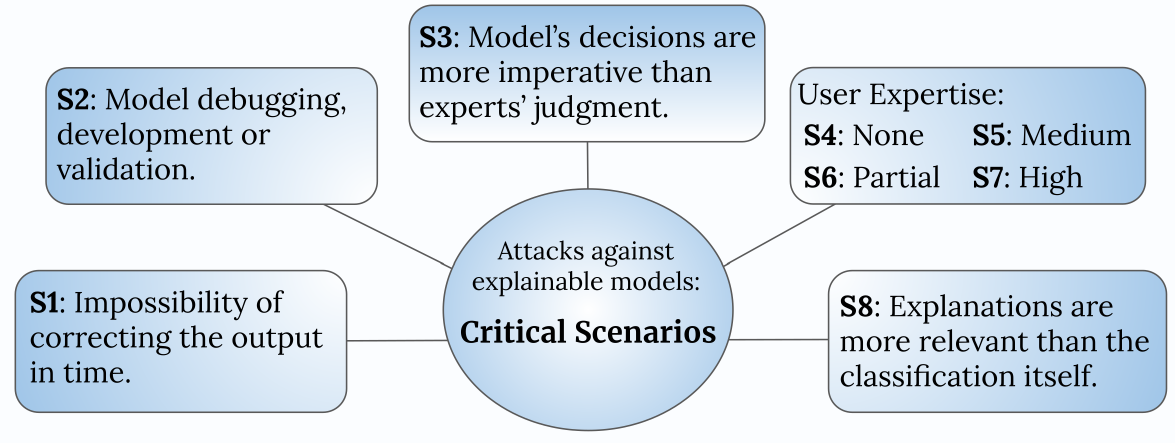}
    \caption{\mfrr Critical scenarios to be considered in the study adversarial attacks against explainable machine learning models.}
    \label{fig:scenarios}
\end{figure}

{\mfrr The considered scenarios, summarized in Figure \ref{fig:scenarios},} comprise different degrees of expertise of the human in supervision of the classification process and different purposes of the explanation.
It is important to note that a particular problem or task could belong to more than one scenario. Moreover, we emphasize that some of the scenarios involve factors which are difficult to quantify in a formal way (e.g., the expertise of a user). Nevertheless, we believe that it is necessary to consider such detailed scenarios in order to rigorously discuss which type of adversarial examples can be realizable in practice.
{\mfrr In what follows, we describe each scenario and identify the requirements that adversarial attacks should satisfy in order to pose a realistic threat in each of them. This information will be summarized in Table \ref{tab:scenarios}.}

\begin{table}[]
\centering
\scalebox{0.8}{
\begin{tabular}{|l|l|l|}
\hline
Scenario &
  Representative Examples &
  Applicable Attacks
  \\
  \hline

\begin{tabular}[c]{@{}l@{}}
\textbf{S1:} 
Impossibility of \\
correcting the output \\
or \ controlling \ the \\
implications of the \\
decision in time.
\end{tabular} 
  & 
  \begin{tabular}[c]{@{}l@{}}
  Fast decision making \\
  scenarios \ \ (e.g., \ \ au-\\ 
  tonomous \ \ cars) \ \ or \\
  automatized  processes \\ 
  (e.g., \ \ massive \ online \\
  content filtering).
  \end{tabular} 
&
  \begin{tabular}[c]{@{}l@{}}
  $\bullet$ 
  Any adversarial attack capable of\\
  producing \ a \ change \ in \ the output \\
  class (as \ the \ explanations \ are \ not \\
  of practical use in these cases).
  \end{tabular} 
  \\ 
  \hline

\begin{tabular}[c]{@{}l@{}}
\vspace{0.1cm} \\
\textbf{S2:} 
Model debugging, \\
development, \\
validation, etc. \\
\vspace{0.1cm}
\end{tabular} 
   &
  \begin{tabular}[c]{@{}l@{}} 
  Applicable to any task.
  \end{tabular}
&
\multirow{2}{*}{
  \begin{tabular}[c]{@{}l@{}}
  $\bullet$
  A.2.1, A.3 \ (justify misclassifica-\\
  tions of the model).  \\
  $\bullet$
  A.1.1 \ (mask \  inappropriate \ be-\\ 
  haviors, \ \  e.g., \ \ hiding \ \ biases by \\
  producing trustworthy outputs or\\
  explanations). \\
  $\bullet$
  A.2 \ \ (produce \ wrong \ outputs \\
  and \ explanations \ jointly).
   \end{tabular} 
  }
  \\ 
  \cline{1-2}

\begin{tabular}[c]{@{}l@{}}
\vspace{0.1cm} \\
\textbf{S3:} 
Decisions \ of \ the \\ 
models \ are \ more \ im-\\
perative  than experts' \\
judgments. \\
\vspace{0.1cm}
\end{tabular} 
  &
  \begin{tabular}[c]{@{}l@{}} 
  Risk \ of \ criminal \\
  recidivism or credit \\
  risk \ management.
  \end{tabular}
&
  \\
  \hline
  
\begin{tabular}[c]{@{}l@{}}
\textbf{S4:} 
User with no\\
expertise.
\end{tabular} 
& 
  \begin{tabular}[c]{@{}l@{}}
  Scenarios in which the \\
  decision \ \ \ criteria \ \ are \\
  secret, \ \ \ \ hidden, \ \ \ \ or \\
  unknown (e.g., banking \\
  or \ financial \ \ scenarios, \\
  malware \ \ classification  \\
  problems, etc.). 
  \end{tabular}
  &
  \begin{tabular}[c]{@{}l@{}}
  $\bullet$ 
  Any \ adversarial \, attack \ scheme \\
  (able \ to \ change \ the \ classification, \\
  the \, explanation \, or \, both \, at \, the \\
  same time), taking advantage of the \\
  user's inexperience.
  \end{tabular} 
   \\ 
   \hline

\begin{tabular}[c]{@{}l@{}}
\textbf{S5:}
User with medium \\
expertise \ \ (the \ \ model \\
is \ expected \ to \ clarify \\
or \ support \ the \ user's \\
decisions).  
\end{tabular} 
  & 
  \begin{tabular}[c]{@{}l@{}}
  Challenging \ scenarios \\ 
  (e.g., complex medical  \\
  diagnosis) \  or \ unfore- \\ 
  seeable scenarios (e.g., \\ 
  macroeconomic predic-  \\
  tions, \ risk \ of \ criminal \\
  recidivism, etc.).
  \end{tabular}
&
  \begin{tabular}[c]{@{}l@{}}
  $\bullet$ 
  A.1.1, A.2.1, A.3. \\ 
  $\bullet$ 
  The \ explanation \ needs \ to \ be \\ 
  consistent with the input patterns  \\
  and/or consistent with the output \\
  class. 
  \end{tabular}
  \\
  \hline

 \begin{tabular}[c]{@{}l@{}}
 \textbf{S6:} 
 User with partial \\
 expertise (i.e., expert \\
 in \ some \ factors \ but \\
 clueless in others).
 \end{tabular} 
     &
  \begin{tabular}[c]{@{}l@{}}
  Hierarchical classification \\
  (e.g., \ \ large \ scale \ visual \\
  recognition).
  \end{tabular}
   &
  \begin{tabular}[c]{@{}l@{}}
  $\bullet$ 
  A.1.1, A.2.1, A.3. \\ 
  $\bullet$ 
  The \  output \ and \ \ the  \ \ explanation \\
  should be consistent \ with the factors \\
  which are familiar to the user (either \\ 
  regarding input features or the output \\
  class).
  \end{tabular} 
  \\ 
  \hline

\begin{tabular}[c]{@{}l@{}}
\textbf{S7:} User \ with \ high \\
expertise.
\end{tabular} 
  &
  \begin{tabular}[c]{@{}l@{}}
  Tasks in which the inputs \\
  can be ambiguous \ (e.g.,   \\
  NLP tasks \ such as senti-        \\
  ment analysis or multiple \\
  object \ detection \ in \ the \\
  image domain).
  \end{tabular}
&
  \begin{tabular}[c]{@{}l@{}}
  $\bullet$ 
  A.1.1, \ A.3 \ (attacks \ involving \\
  generating ambiguous explanations).
  \end{tabular} 
  \\ 
  \hline

\begin{tabular}[c]{@{}l@{}}
\textbf{S8:} 
Explanations even \\
more relevant than the \\
classification \ itself. 
\end{tabular} 
  &
  \begin{tabular}[c]{@{}l@{}}
  Predictive maintenance, \\
  medical \ diagnosis \ or \\
  credit/loan \ \ approval \\
  (e.g., \ \ with \ a \ wrong  \\
  explanation users can \\
  not modify or correct \\
  the deficiencies).  \\
  \end{tabular}
  &
  \begin{tabular}[c]{@{}l@{}}
  $\bullet$ 
  A.1, A.2.1 (e.g., maintain the output \\ 
  but produce totally or partially wrong \\
  explanations, \ or \ produce \ unethical \\
  explanations).
  \end{tabular} 
  \\
  \hline
  
\end{tabular}
}
\caption{Possible scenarios in which explainable machine learning models can be deployed, and a guideline on how adversarial attacks should be designed in each case in order to pose a realistic threat.}
\label{tab:scenarios}
\end{table}

%scenarios_summary

\paragraph{S1 Scenario:} The first scenario comprises tasks in which the implications of the decisions made by the model cannot be controlled by the user, or cases in which there is no time for human supervision of the predictions. 
{\mfrr Despite the relevance of some tasks that fall into this category, such as autonomous driving \citep{fujiyoshi2019deep} or massive content filtering \citep{mahdavifar2019application,kuchipudi2020adversarial}, humans cannot thoroughly evaluate each possible prediction.}
For this reason, explanations are not of practical use in such cases, so the main (or only) goal of an adversary is to produce an incorrect output.

\paragraph{S2 Scenario:} 
{\mfrr
Interpretability or explainability can be desirable properties for machine learning models (including those developed for the \textbf{S1} scenario) in order to \textbf{debug or validate} them \citep{ras2018explanation,doshi-velez2018considerations,adadi2018peeking,anders2022finding}.}
{\mfrr
For instance, a model developer might want to explain the decisions of a self-driving car (even if the end-user will not receive explanations when the model is put into practice) to assess why it has provided an incorrect output, to validate its reasoning process or to gain knowledge about what the model has learned \citep{ras2018explanation,fujiyoshi2019deep,mori2019visual}.} In such cases, an adversary could: justify a misclassification of the model (A.2.1, A.3), hide an inappropriate behavior when the model predicts correctly but for the wrong reasons (A.1.1), or produce wrong outputs and explanations at the same time (A.2). 

\paragraph{S3 Scenario:}
The same attack strategies {\mfrr applicable to the \textbf{S2} scenario} 
can be applied in scenarios in which the models' decisions are taken as more relevant or imperative than the experts' judgments. {\mfrr Although this scenario resembles S1,} the main difference is that, in this case, explanations can be useful or relevant even when the model is deployed or employed by the end-user, and, therefore, the attack should also take the explanations into consideration instead of considering only the output class.

\paragraph{S4 Scenario:} Regarding the expertise level of the user querying the model, the case of no expertise is the simplest one from the perspective of the adversary,  as any attack scheme can be produced without arousing suspicions, taking advantage of the user inexperience. For the same reason, models deployed in such scenarios should also be the ones with more security measures against adversarial attacks.

\paragraph{S5 Scenario:}
If the user's expertise is medium, the model might be expected to clarify or support the user's decisions. Thus, the explanation should be sufficiently consistent with the main semantic features in the input 
{\mfrr
(e.g., the user might not be able to diagnose a medical image, but can identify the relevant spots depending on what is being diagnosed, such as darker spots in skin-cancer diagnosis \citep{al-masni2018skin}),
}
and/or be sufficiently consistent with the output class (A.1.1, A.2.1, A.3). 

\paragraph{S6 Scenario:}
Similarly to the \textbf{S5} scenario, if the user has a partial expertise, that is, if the user is an expert in some factors but clueless in others,\footnote{\mfrr This could happen in hierarchical classification tasks or large-scale visual recognition tasks, as a fine-grained distinction of certain classes might be challenging, whereas the remaining classes are easily classified \citep{silla2011survey,deng2009imagenet,russakovsky2015imagenet,hase2019interpretable,nguyen2021effectiveness}.}
then the adversary needs to ensure that the output and the explanations are coherent only with the factors or features that are familiar to the user {\mfrr (A.1.1, A.2.1, A.3)}.

\paragraph{S7 Scenario:}
A user with high expertise, by definition, will realize that a model is producing a wrong output or explanation. However, it can be possible to mislead the model and convince the human of a wrong prediction by means of ambiguity (A.1.1, A.3).
{\mfrr
For instance, in an image classification task, two objects can appear at the same time, making it possible to produce a wrong class with a reasonable explanation, for example, by selectively focusing the attention of the explanation on one of the objects or by highlighting the secondary object as the most relevant one \citep{stock2018convnets}.}
{\mfrr
In addition, in problems in which the inputs are inherently ambiguous, such as natural language processing tasks, different but reasonable explanations can be produced for the same input \citep{agirre2006word,beck2020representation}.}

\paragraph{S8 Scenario:}
Finally, in some cases the explanations might be more critical, necessary or challenging than the output itself. Some representative tasks are {\mfrr predictive maintenance \citep{serradilla2022deep}
(e.g., it might be more interesting to know why a certain system will fail than just knowing that it will fail) or medical diagnosis \citep{stiglic2020interpretability} (e.g., discovering why a model has diagnosed a patient as being at high risk for a particular disease might be the main priority to prevent the disease or provide a better treatment).} For these reasons, a change in the explanation is critical for such models, which makes them particularly sensitive to the attacks described in A.1, or those described in A.2.1, if the misclassification of the model is difficult to notice by the user.

\section{Illustration of context-aware adversarial attacks}
\label{sec:experiments}

In this section we generate different types of adversarial examples to illustrate the main attack paradigms described in Section \ref{sec:adv_for_explainable_ml}, in terms of both the type of misclassification that wants to be produced (as described in Section  \ref{sec:explanation_aware}) and the ``scenario'' in which the attack is created (as described in Section \ref{sec:scenarios}). To this end, we will consider two representative image classification tasks, assuming an explainable machine learning scenario. In addition, we will consider two explanation methods, namely feature-based explanations and prototype-based explanations, to illustrate the effect of the attacks in both cases. Our code is publicly available at: \url{https://github.com/vadel/AE4XAI}.

{\mfrr We remark that the aim of this section is to provide illustrative examples of the attack paradigms described in the proposed framework, and that the focus will be on exemplifying the design of the attacks (i.e., the requirements that they should satisfy in order to pose a legit threat against explainable models) rather than on the methods that could be used to implement them or in their performance. A summary of the illustrated scenarios and the corresponding details is provided in Table \ref{tab:experiment_summary}. As can be seen in the table, our illustrations cover all the main attack paradigms and scenarios considered in the framework developed in the previous section.
}

\subsection{Selected tasks, datasets and models}
\label{sec:tasks_datasets_models}

We will focus on two image classification tasks to generate the adversarial examples:

\begin{itemize}
\item \emph{Medical image classification:} the selected task consists of chest X-ray (CXR) classification, in which the aim is to identify, given an X-ray image, one of the following diseases: Covid-19, (non-Covid) pneumonia or none (``normal''). We used a pretrained Covid-Net model \citep{wang2020covidnet}, trained on the COVIDx dataset \citep{wang2020covidnet}, which achieves an accuracy of 92.6\%.\footnote{The selected pretrained model (COVIDNet-CXR Small) is accessible at \url{https://github.com/lindawangg/COVID-Net/blob/master/docs/models.md}.}

\item \emph{Large scale visual recognition:} the aim of this task is to classify \textit{real}  or \textit{natural} images across a wide range of classes. We selected the ImageNet dataset, which contains images from 1000 different classes such as animals or ordinary objects, and a pre-trained ResNet-50 deep neural network as a classifier, which achieves a Top-1 accuracy of 74.9\%.\footnote{More information about the pretrained model used can be found at \url{https://keras.io/api/applications/resnet/}.} Both the ImageNet and the ResNet-50 architecture have been widely employed for the study of image classification, as well as in the more particular field of interpretable machine learning \citep{selvaraju2017gradcam, zhang2020interpretable,  nguyen2021effectiveness}.
\end{itemize}
These two use-cases allow us to illustrate the different scenarios described in Section \ref{sec:scenarios}. First, the medical image classification represents a challenging task that requires a high expertise in order to correctly classify inputs or to provide well-funded explanations of the decisions. As discussed, in such a scenario, an adversary has more room to generate adversarial examples that produce incorrect model responses (both in terms of classification and explanation) which, at the same time, may be coherent or acceptable according to a human criterion (particularly for non-expert users). Moreover, the explanations can be critical in this task, as the reason for determining a diagnosis is of high relevance, being therefore representative of the S8 scenario described in Section \ref{sec:scenarios}.

Secondly,  users with a high expertise can be assumed in the large scale visual recognition task, as the ImageNet dataset contains images containing familiar objects or animals which will be easily recognizable for humans. Thus, a human observing the input as well as the output of the classification should easily detect inconsistencies in the prediction of the model (i.e., whether or not it is correct). However, at the same time, some images might be ambiguous or challenging to classify even for humans (e.g., fine-grained dog breed classification \citep{nguyen2021effectiveness,khosla2011novel}) which therefore can be representative of medium or partial expertise, as the user might be able to effectively discriminate certain classes (e.g., differentiating dogs from other animal species) but not others (e.g., two similar dog breeds). 
In such cases, the user might expect the prediction of the model or the corresponding explanation to clarify the correct class of the input.

\begin{table}[]
\centering
\scalebox{0.8}{
{\mfrr
\begin{tabular}{|l|c|c|c|c|c|c|}
\hline 
\multicolumn{1}{|c|}{Task} &
  \begin{tabular}[c]{@{}c@{}}Type of \\ Explanation\end{tabular} &
  \begin{tabular}[c]{@{}c@{}}Possible \\ Scenario\end{tabular} &
  \begin{tabular}[c]{@{}c@{}}Wrong\\ class\end{tabular} &
  \begin{tabular}[c]{@{}c@{}}Wrong\\ explanation\end{tabular} &
  \begin{tabular}[c]{@{}c@{}}Attack \\ description\end{tabular} &
  Figure \\ 
  \hline
  \hline
\multicolumn{1}{|c|}{\multirow{6}{*}{\begin{tabular}[c]{@{}l@{}}X-ray\\ (Sec. \ref{sec:attacks_xray}) \end{tabular}}} &
  \multirow{6}{*}{\begin{tabular}[c]{@{}c@{}}Feature-based\\ (saliency map)\end{tabular}} &
  \multirow{6}{*}{\begin{tabular}[c]{@{}c@{}}S2, S3, \\ S4/S5/S6,\\ S8\end{tabular}} &
  \xmark &
  \xmark &
  \begin{tabular}[c]{@{}c@{}}No attack\\ (i.e., original input)\end{tabular} &
  \ref{fig:x_ray_adversarials}-(a) \\ \cline{4-7} 
\multicolumn{1}{|c|}{} &
   &
   &
  \cmark &
  \begin{tabular}[c]{@{}c@{}}\cmark\\ (conflicting)\end{tabular} &
   \begin{tabular}[c]{@{}c@{}}Regular attack\\ (i.e., without control-\\ ling the explanation)\end{tabular} &
  \ref{fig:x_ray_adversarials}-(b) \\ \cline{4-7} 
\multicolumn{1}{|c|}{} &
   &
   &
  \cmark &
  \begin{tabular}[c]{@{}c@{}}\xmark\\ (non-conflicting)\end{tabular} &
  A.3 &
  \ref{fig:x_ray_adversarials}-(c) \\ \cline{4-7} 
\multicolumn{1}{|c|}{} &
   &
   &
  \xmark &
  \begin{tabular}[c]{@{}c@{}}\cmark\end{tabular} &
  \begin{tabular}[c]{@{}c@{}}A.1.1\\ (confusing \\ recommendation)\end{tabular} &
  \ref{fig:x_ray_adversarials}-(d) \\ \cline{4-7} 
\multicolumn{1}{|c|}{} &
   &
   &
  \cmark &
  \begin{tabular}[c]{@{}c@{}}\cmark\\ $A_f(x) \sim f(x)$\\ (non-informative \\ but consistent, i.e.,\\ supports prediction)\end{tabular} &
  A.2.1 &
  \ref{fig:x_ray_adversarials}-(e) \\ \cline{4-7} 
\multicolumn{1}{|c|}{} &
   &
   &
  \cmark &
  \begin{tabular}[c]{@{}c@{}}\cmark\\ $A_f(x) \not\sim f(x)$\\ $A_f(x) \not\sim y_x$ \ \ \ \\  (non-informative\\ and inconsistent)\end{tabular} &
  A.2.2 &
  \ref{fig:x_ray_adversarials}-(f) \\ 
  \hline
  \hline
\multirow{7}{*}{\begin{tabular}[c]{@{}l@{}}Large-\\ Scale\\ Visual\\ Recog. \\ (Sec. \ref{sec:attacks_lsvr}) \end{tabular}} &
  \multirow{6}{*}{\begin{tabular}[c]{@{}c@{}}Feature-based\\ (saliency map)\end{tabular}} &
  \multirow{2}{*}{\begin{tabular}[c]{@{}c@{}}S2,\\ S5/S6\end{tabular}} &
  \xmark &
  \xmark &
  \begin{tabular}[c]{@{}c@{}}No attack\\ (i.e., original input)\end{tabular} &
  \ref{fig:imagenet_dogs_adv}-(a) \\ \cline{4-7} 
 &
   &
   &
  \cmark &
  \xmark &
  A.3 &
  \begin{tabular}[c]{@{}c@{}}
  \ref{fig:imagenet_dogs_adv}-(b),\\
  \ref{fig:imagenet_dogs_adv}-(c),\\
  \ref{fig:imagenet_dogs_adv}-(d) \!\end{tabular} \\ \cline{3-7} 
 &
   &
  \multirow{4}{*}{S2, S7} &
  \xmark &
  \xmark &
  \begin{tabular}[c]{@{}c@{}}No attack\\ (i.e., original input)\end{tabular} &
  \ref{fig:imagenet_dog_suit_adv}-(a) \\ \cline{4-7} 
 &
   &
   &
  \xmark &
  \xmark &
  \begin{tabular}[c]{@{}c@{}}No attack\\ (the output is further\\ biased in favour of the \\ correct class, avoiding \\ ambiguities)\end{tabular} &
  \ref{fig:imagenet_dog_suit_adv}-(b) \\ \cline{4-7} 
 &
   &
   &
  \cmark &
  \cmark &
  \begin{tabular}[c]{@{}c@{}}A.2.1\end{tabular} &
  \ref{fig:imagenet_dog_suit_adv}-(c) \\ \cline{4-7} 
 &
   &
   &
  \cmark &
  \xmark &
  \begin{tabular}[c]{@{}c@{}}A.3\end{tabular} &
  \ref{fig:imagenet_dog_suit_adv}-(d) \\ \cline{2-7} 
 &
  \begin{tabular}[c]{@{}c@{}}Prototype-based\\ explanation\\ (3 nearest \\ training inputs)\end{tabular} &
  \begin{tabular}[c]{@{}c@{}}S2, \\ S5/S6\end{tabular} &
  \cmark &
  \begin{tabular}[c]{@{}c@{}} \xmark \\  $A_f(x)\sim y_x$ \\ $A_f(x)\sim f(x)$ \\ (ambiguous) \end{tabular} &
  A.3 &
  \begin{tabular}[c]{@{}c@{}}
  \ref{fig:adv_proto_exp}-(a),\\ 
  \ref{fig:adv_proto_exp}-(b) \! \end{tabular} \\ \hline
\end{tabular}
}
}
\caption{\mfrr Summary of the illustrative attacks shown in Sections \ref{sec:attacks_xray} and \ref{sec:attacks_lsvr}. Notice that each attack paradigm and scenario is exemplified at least once. Note also that for the large-scale visual recognition task different scenarios can be considered depending on the characteristics or the challengingness of the input.}
\label{tab:experiment_summary}
\end{table}

\subsection{Explanation methods}

We will consider two representative explanation methods in order to illustrate an explainable machine learning scenario:

\paragraph{Feature-based explanation}
The Grad-CAM method \citep{selvaraju2017gradcam} will be used to generate saliency-map explanations. The rationale of this method is to employ the feature maps learned by the model in the last convolutional layer to produce the explanations. 
Given a convolutional neural network $f$ and an input $x$, the Grad-CAM saliency-map $S$ is defined as:
\begin{equation}
\label{eq:gradcam}
    S = ReLU \left( \sum_{m}^M \alpha_{m,c}\cdot C_m \right),
\end{equation}
where $C_{m}$, $m=1,\dots,M$, represents the (two-dimensional) $m$-th activation map (for the input $x$) at the last convolutional layer of $f$, and  $\alpha_{m,c}\in \mathbb{R}$ represents the importance of the $m$-th map in the prediction of the class of interest $y_c$ (typically $f(x)$, i.e., the class predicted by the model). The importance $\alpha_{m,c}$ of each activation map is estimated as the average global pooling of the gradient of the output score (corresponding to the class $y_c$) with respect to $C_m$, which will be denoted as $G_{m,c} = \nabla_{C_m}f_c(x)$:
\begin{equation}
    \alpha_{m,c} = \sum_{i} \sum_{j} G_{m,c}^{i,j},
\end{equation}
where $G_{m,c}^{i,j}$ denotes the value at the $i$-the row and $j$-th column. The ReLU non-linearity in \eqref{eq:gradcam} is applied to remove negative values, maintaining only the features with a positive influence on $y_c$.\\

\paragraph{Example-based explanation}
We will also consider an example-based explanation in which the $k$ training images (which can be considered prototypes representing classes) that are closest to the input which has been classified are provided \citep{nguyen2021effectiveness}. The proximity between the inputs will be measured as the Euclidean distance of the $l$-dimensional latent representation $r_f(x): \mathbb{R}^d \rightarrow \mathbb{R}^l$ learned by the model $f$ in the last layer, that is, the (flattened) activations of the last convolutional layer of the model. 
This representation captures complex semantic features of the inputs, thus providing a more appropriate representation space for meaningfully comparing input samples according to the features learned by the model. Let $X^c_{train}$ represent the set of \textit{training} inputs belonging to the class of interest $y_c$ (e.g., the class predicted by the model). Given a model $f$ and an input $x$, the explanation will be a set $P \subseteq  X^c_{train}$, with $|P|=k$, that satisfies:

\begin{equation}
||r_f(\hat{x})-r_f(x)||_2 \ > \ || r_f(x^p)-r_f(x)||_2, \ \ \forall \hat{x} \in X^c_{train} \! - \! P, \ \forall x^p \in P.
\end{equation}

Note that the two selected methods allow, by definition, explanations to be computed for any class of interest $y_c$. However, we will consider as the \textit{main} explanation the one corresponding to the predicted class $f(x)$. 
{\mfrr Finally, we assume that the explanation methods and their parameters are fixed and known to the adversary. Since the focus of our experimentation is illustrative and not performance-based, analyzing the sensitivity of the explanation methods to hyperparameters \citep{bansal2020sam,dombrowski2019explanations} will be out of the scope of this section.}

\subsection{Attack method}

We will assume a targeted attack for our experiments, in which the aim will be to create, given an input $x$, an adversarial example $x'$ such that:
\begin{align}
 & f(x')  \ \, \,   = y_t,   \\
 & A_f(x')   = \xi_t,   \\
 & ||x-x'||  \leq \epsilon,
\end{align}
where $y_t$ represents a target class, $\xi_t$ a target explanation and $\epsilon$ the maximum distortion norm. For the case of saliency-map explanations, $\xi_t$ will be a predefined saliency-map $S_t$. For the case of prototype-based classification, $\xi_t$ will be the set $P_t$ of $k$ training inputs (with the value of $k$ fixed beforehand by the explanation method) selected by the adversary to be produced as explanations (that is, the training inputs of class $y_t$ that are closer to $x$ should be those in the set $P_t$). We do not specify any particular order for the $k$ target-prototypes in $P_t$, that is, we assume that the relevance of each of the $k$ prototypes in the explanation is the same.

We will use a targeted Projected Gradient Descent (PGD) attack \citep{madry2018deep} to generate the adversarial examples. This attack iteratively perturbs the input sample in the direction of the gradient of a loss function $L$ (e.g., the cross-entropy) with respect to the input, $\mathrm{sign}(\nabla_{x'_i} L(x'_i,y_t))$, with a step size $\alpha$. At each step, the adversarial example is projected {\mfrr by a projection operator $\mathcal{B}_{\epsilon}$} to ensure that the norm of the adversarial perturbation $v=x'-x$ is restricted to $||v||_\infty \leq \epsilon$:
\begin{equation}
\label{eq:pgd}
x'_{i+1} = \mathcal{B}_{\epsilon}(x'_i - \alpha \cdot \mathrm{sign}(\nabla_{x'_i} L(x'_i,y_t)) ).
\end{equation}
In order to produce attacks capable of changing both the classification and the explanation, we will consider a generalized loss function
\begin{equation}
\label{eq:loss_pred_exp}
L(x_i,y_t,\xi_t,\lambda) = (1-\lambda)\cdot L_{pred}(x_i,y_t) + \lambda \cdot L_{expl}(x_i,\xi_t),
\end{equation}
where $L_{pred}(x,y_t)$ represents the classification error with respect to the target class $y_t$, $L_{expl}(x,\xi_t)$ represents the explanation error with respect to the target explanation $\xi_t$ and $\lambda \in [0,1]$ balances the trade-off between both functions. A close approach can be consulted in \citet{zhang2020interpretable}. In our experiments, we used the cross-entropy loss as $L_{pred}$. For the case of saliency-map explanations, we instantiated $L_{expl}$ as the Euclidean distance between the model's explanation $g(x)=S$ and the target saliency map $S_t$ (specified by the adversary):
\begin{equation}
L_{expl}(x,S_t)=||g(x) - S_t||_2.
\end{equation}
For the case of prototype-based explanations, $L_{expl}$ will be the average Euclidean distance between the latent representation of the (adversarial) input and the latent representation of the $k$ prototypes selected by the adversary as the target explanation $P_t$:
\begin{equation}
L_{expl}(x,P_t)=\frac{1}{k}\sum_{x^p \in P_t} ||r(x)-r(x^p)||_2.
\end{equation}

\subsection{Illustrative attacks in the X-ray classification task}
\label{sec:attacks_xray}

\begin{figure}[!b]
    \centering
    \subfloat[]{\includegraphics[scale=0.49]{original_xray-eps-converted-to.pdf}
    \hskip -9pt
    \includegraphics[scale=0.49]{original_xray_exp-eps-converted-to.pdf}}
    \hskip 7pt
    \subfloat[]{\includegraphics[scale=0.49]{adv_reg_xray-eps-converted-to.pdf}
    \hskip -9pt
    \includegraphics[scale=0.49]{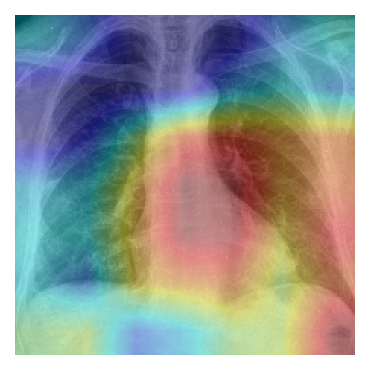}}  
    \\
    \subfloat[]{\includegraphics[scale=0.49]{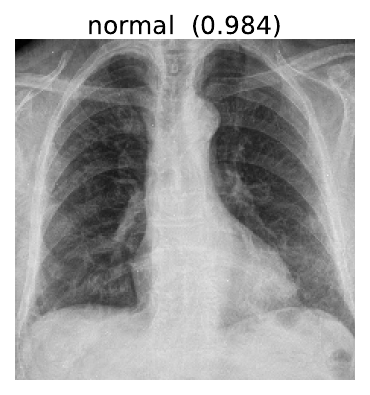}
    \hskip -9pt
    \includegraphics[scale=0.49]{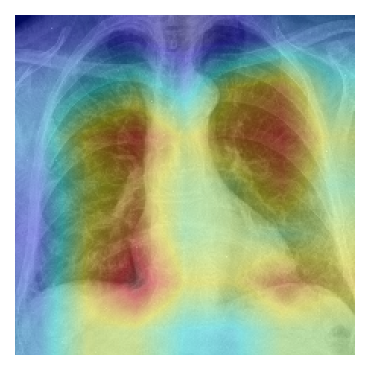}}
    \hskip 7pt
    \subfloat[]{\includegraphics[scale=0.49]{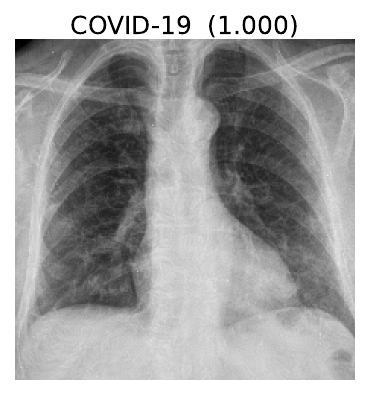}
    \hskip -9pt
    \includegraphics[scale=0.49]{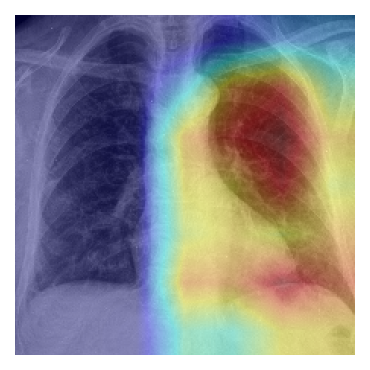}}
    \\
    \subfloat[]{\includegraphics[scale=0.49]{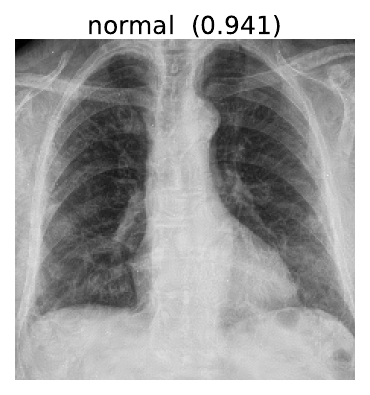}
    \hskip -9pt
    \includegraphics[scale=0.49]{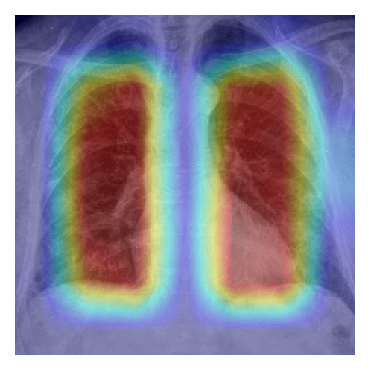}}
    \hskip 7pt
    \subfloat[]{\includegraphics[scale=0.49]{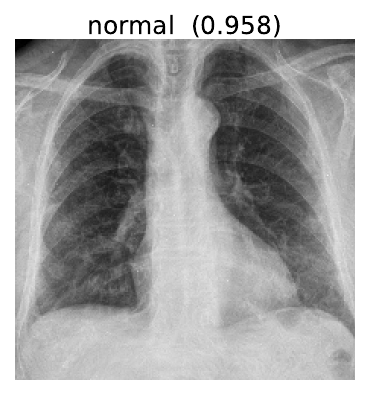}  
    \hskip -9pt
    \includegraphics[scale=0.49]{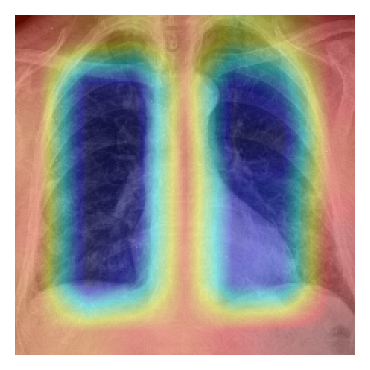}}
%\end{tabular}
    \caption{Different types of adversarial attacks for the x-ray medical image diagnosis task. The left part of each image shows the input image as well as the class assigned by the model (jointly with the confidence score in the $[0,1]$ range), whereas the right part shows the explanation provided by the Grad-CAM method. %\\ 
    (a) Original image. %\\
    (b) Regular adversarial attack (PGD) targeting the class ``normal'' (i.e., the possible changes that the adversarial perturbation may produce in the explanation are not controlled by the attack). %\\
    (c) Attack producing the wrong classification ``normal'' while maintaining the original explanation. %\\
    (d) Attack maintaining the correct classification while changing the explanation in order to selectively highlight some parts (the right part) but omitting others (in this case, the left part). %\\
    (e) Attack producing the wrong class ``normal'' and a wrong explanation which uniformly highlights the relevant parts of the image. %\\
    (f) Attack producing the wrong class ``normal'' and a uniform explanation outside the main parts of the image (i.e., highlighting only irrelevant and incorrect parts).}
    \label{fig:x_ray_adversarials}
\end{figure}

Figure \ref{fig:x_ray_adversarials} illustrates the results obtained for different adversarial examples generated against the COVID-Net model. The left part of each sub-figure shows the input sample, the model's classification of the input and the confidence score of the prediction, whereas the right part shows the saliency-maps generated with the Grad-CAM explanation (darker-red parts represent a higher relevance). Figure \ref{fig:x_ray_adversarials}-(a) shows the original (i.e., unperturbed) input sample, which is correctly classified as its ground-truth class ``COVID-19''.\footnote{Disclaimer: the authors acknowledge no expertise in CXR classification, and, as the dataset does not contain a ground-truth saliency-map explanation, it will be assumed for illustration purposes that the explanation achieved for the original input is coherent and correct.}

Figure \ref{fig:x_ray_adversarials}-(b) shows an adversarial example generated using a regular PGD attack (considering only changing the output class, i.e., $L=L_{pred}$), targeting the class ``normal''. Notice that the main parts of the explanation have changed to the central part and to the rightmost part of the image, highlighting mainly irrelevant zones. Therefore, such an explanation might not be taken as consistent. Contrarily, Figure \ref{fig:x_ray_adversarials}-(c) shows an adversarial example using the attack described in Equation \eqref{eq:pgd} using the loss function described in \eqref{eq:loss_pred_exp}, targeting the class ``normal'' and the original saliency-map (i.e., the one obtained for the original input image), illustrating the attack paradigm A.3 described in Section \ref{sec:explanation_aware}.
We clarify that using a regular adversarial attack might not necessarily imply a change in the explanation (or imply that the explanation, even if changed, will necessarily highlight irrelevant zones). Nevertheless, this example allows us to illustrate the need to control the explanation in order to create adversarial examples that are capable of convincing the human that the model's (mis)classifications and the corresponding explanations are coherent or consistent, as discussed in Section \ref{sec:adv_for_explainable_ml}. 

Figure \ref{fig:x_ray_adversarials}-(d) illustrates the attack paradigm A.1.1 described in Section \ref{sec:explanation_aware}, in which the original class is maintained whereas a change in the explanation is produced, in this case, selectively highlighting some regions of the image. Here, we generated the attack setting the target map as the right-half side of the original saliency-map, and setting the left-part values as zero. Such an attack strategy can be extremely concerning for those scenarios in which the explanation is of high relevance, as a misleading adversarial explanation  might lead to an incorrect diagnosis, prescription or treatment. 

Figure \ref{fig:x_ray_adversarials}-(e) illustrates {\mfrr the attack paradigm A.2.1. Notice that} both the output class and the explanation are changed. Moreover, the target map set in this case represents a roughly uniform map over the most relevant parts of the image (in this case the two lungs). Therefore, the provided explanation can be taken as coherent, as the main parts of the image are taken into consideration for the prediction. The fact that the predicted class is ``normal'' also increases the coherence of the explanation, since it can be interpreted from it that the most critical areas are correct (i.e., that there is no evidence in those areas of a possible disease). Indeed, the same explanation would have a different effect if the prediction had represented a disease (e.g., if the original class had been maintained in this case). This is because a uniform explanation would not provide a precise justification of why the disease is predicted, thus hampering a proper diagnosis, but, at the same time, is coherent with the input features (because the most relevant parts are highlighted), contributing to the user's acceptance that the model prediction is correct. Contrarily, in Figure \ref{fig:x_ray_adversarials}-(f) the target map is the opposite: roughly all the relevant parts are considered as not relevant, whereas the remaining regions are considered as relevant, illustrating a case in which the explanation is completely wrong. {\mfrr Since the prediction is also incorrect, and is not supported by the explanation, a total mismatch is produced between all the considered factors, exemplifying the attack paradigm A.2.2.}

%\pagebreak
%\clearpage
%\newpage

\subsection{Illustrative attacks in the large scale visual recognition task}
\label{sec:attacks_lsvr}

\begin{figure}
    \centering
    \subfloat[Original input]{\includegraphics[scale=0.37]{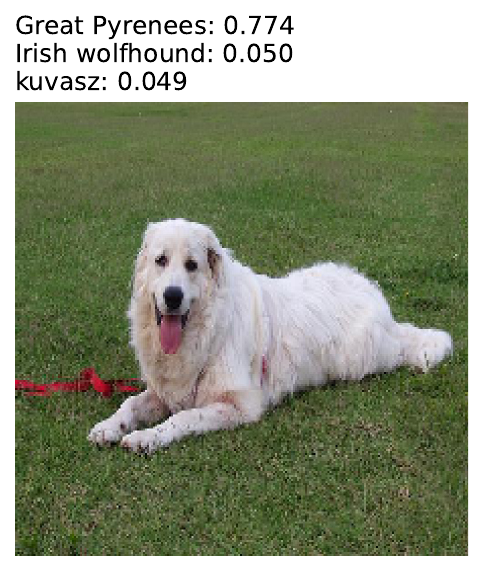}
    \hskip -7pt
    \includegraphics[scale=0.37]{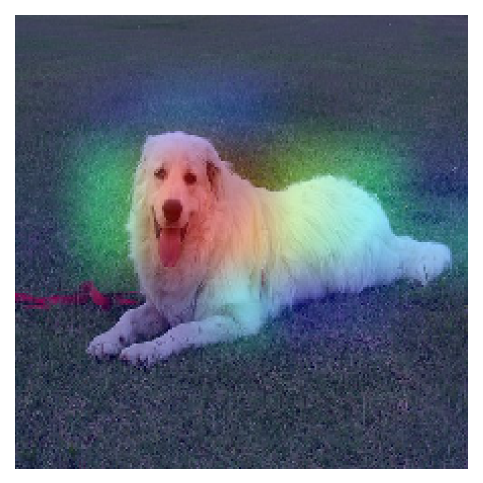}}
    \hskip 6pt
    \subfloat[Target class: Kuvasz]{\includegraphics[scale=0.37]{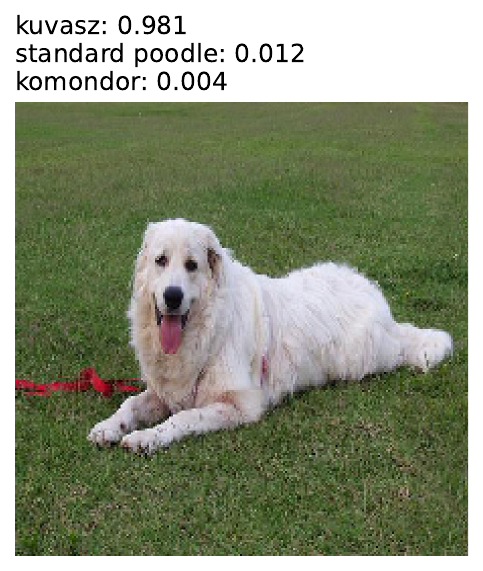}
    \hskip -7pt
    \includegraphics[scale=0.37]{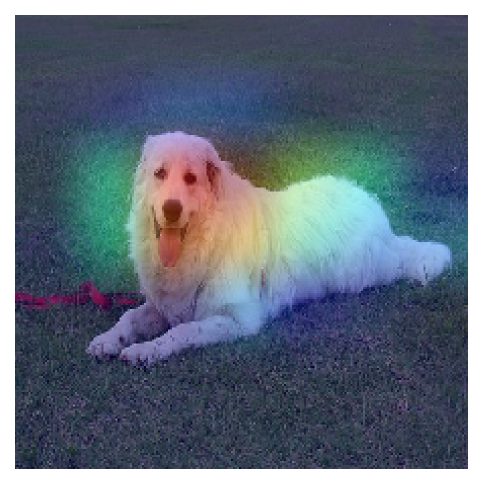}}
    \\
    \subfloat[White wolf]{\includegraphics[scale=0.37]{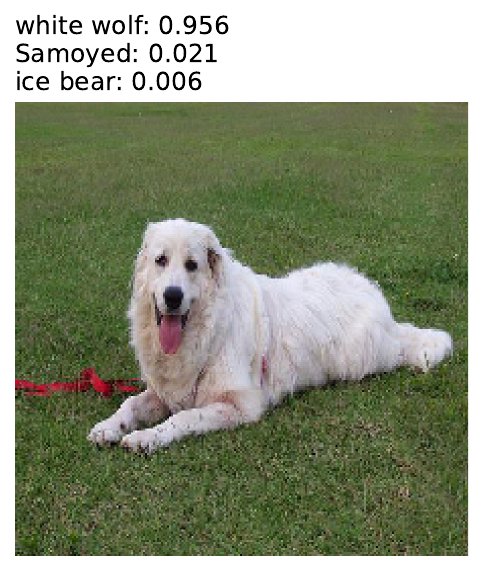}
    \hskip -7pt
    \includegraphics[scale=0.37]{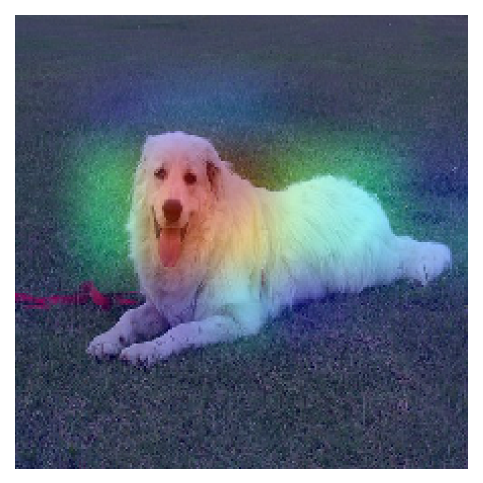}}
    \hskip 6pt
    \subfloat[Labrador Retriever]{\includegraphics[scale=0.37]{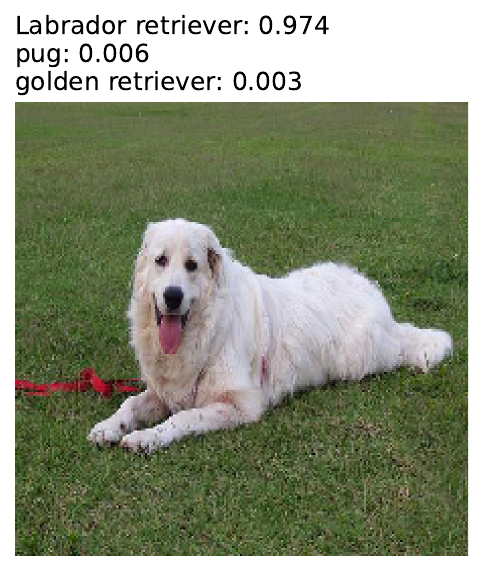}
    \hskip -7pt
    \includegraphics[scale=0.37]{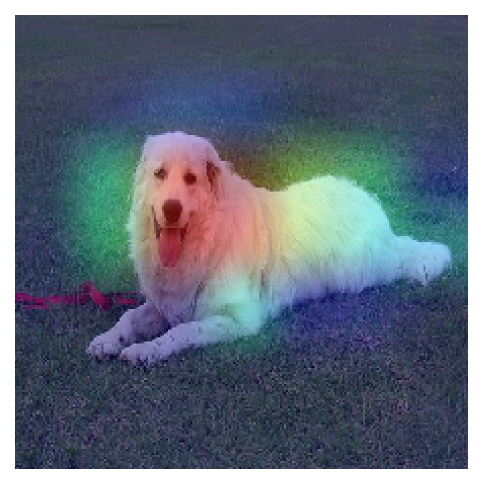}}
   	\\
    \subfloat[Closest ``Great Pyrenees'' inputs]{\includegraphics[scale=0.455]{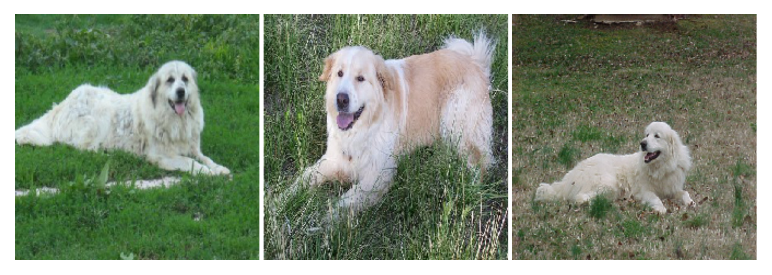}}
    \hskip 5pt
    \subfloat[Closest ``Kuvasz'' inputs]{\includegraphics[scale=0.455]{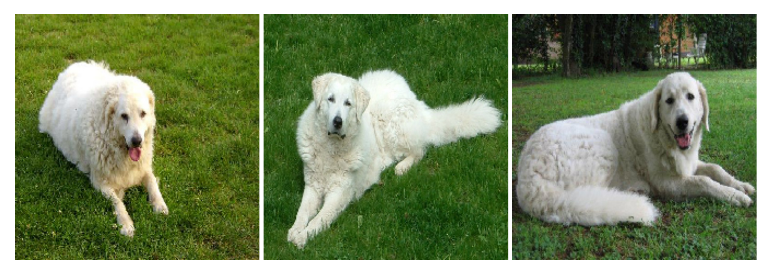}}
    \\
   \subfloat[Closest ``White wolf'' inputs]{\includegraphics[scale=0.455]{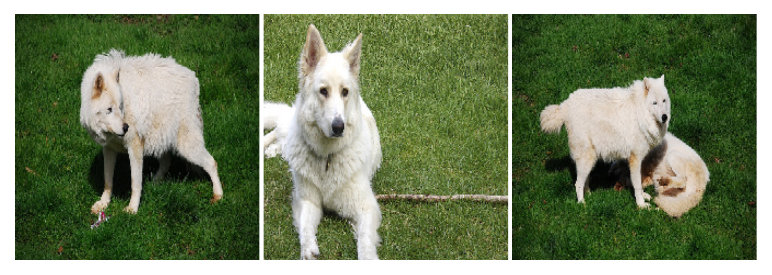}}
    \hskip 5pt
    \subfloat[Closest ``Labrador Retriever'' inputs]{\includegraphics[scale=0.455]{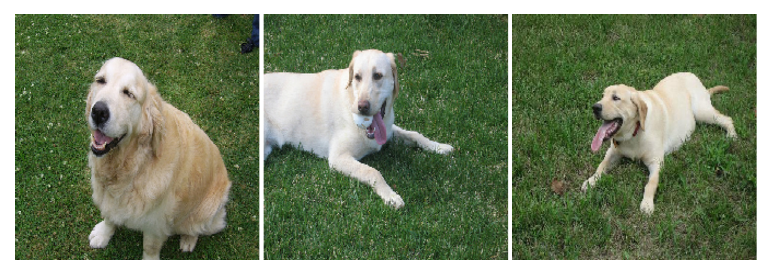}}
    \caption{Adversarial examples generated for the ImageNet dataset classification task taking advantage of class ambiguity. The adversarial examples are generated from the input in (a)-left, which belongs to the source class ``Great Pyrenees'', targeting different classes that are characterized by features similar to those of the source class: (b) ``Kuvasz'', (c) ``White wolf'', and (d) ``Labrador Retriever''. Each adversarial example is created ensuring that the saliency-map explanation of the original input, shown in (a)-right, is maintained. (e)-(h) show, for each of the four classes considered (source class + 3 target classes), the $k=3$ prototypes closest to the original input, in order to assess their similarity.}
    \label{fig:imagenet_dogs_adv}
\end{figure}

In this section, we illustrate different types of adversarial examples generated taking advantage of class ambiguity. First, in Figure \ref{fig:imagenet_dogs_adv}, the similarity between different classes is used to generate adversarial examples capable of producing a misclassification that could be considered as \textit{coherent} or \textit{reasonable} even for humans. These examples illustrate the attack paradigm  A.3 described in Section \ref{sec:adv_for_explainable_ml}. In particular, each attack is generated by setting a target class for which the inputs belonging to that class contain very similar features to those inputs belonging to the source class. Figure \ref{fig:imagenet_dogs_adv}-(a) shows the original input sample used to create the adversarial examples, the top-3 predictions of the model and the corresponding Grad-CAM explanation. Figures  \ref{fig:imagenet_dogs_adv}-(b), \ref{fig:imagenet_dogs_adv}-(c) and \ref{fig:imagenet_dogs_adv}-(d)  show the adversarial examples targeting the classes, ``Kuvasz'', ``White wolf'' and ``Labrador Retriever'', respectively. These classes represent different dog breeds with very similar features, as is shown in Figures \ref{fig:imagenet_dogs_adv}-(e), \ref{fig:imagenet_dogs_adv}-(f), \ref{fig:imagenet_dogs_adv}-(g) and \ref{fig:imagenet_dogs_adv}-(h), in which the prototypes (for each of the classes) that are closer to the original input image are shown. In all the cases, the saliency map targeted in the attack is the one obtained for the original input (i.e., we maintain the original explanation while changing the classification).

\begin{figure}[!h]
    \centering
    \subfloat[Original input]{\includegraphics[scale=0.37]{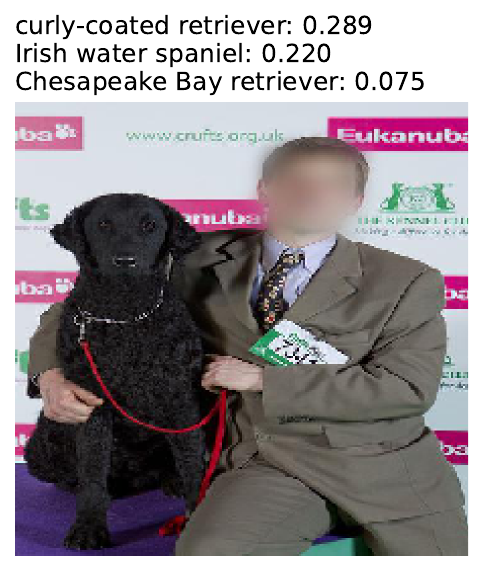}
    \hskip -7pt
    \includegraphics[scale=0.37]{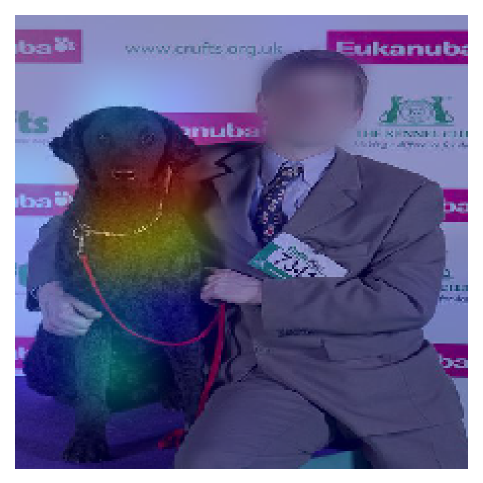}}
    \hskip 6pt
    \subfloat[Target class: ``Curly-coated retriever'']{\includegraphics[scale=0.37]{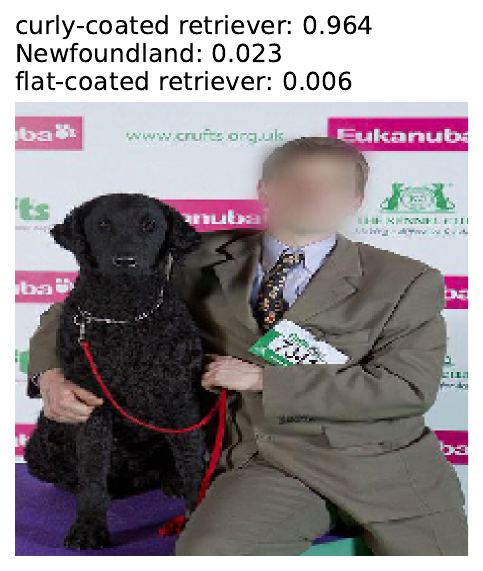}
    \hskip -7pt
    \includegraphics[scale=0.37]{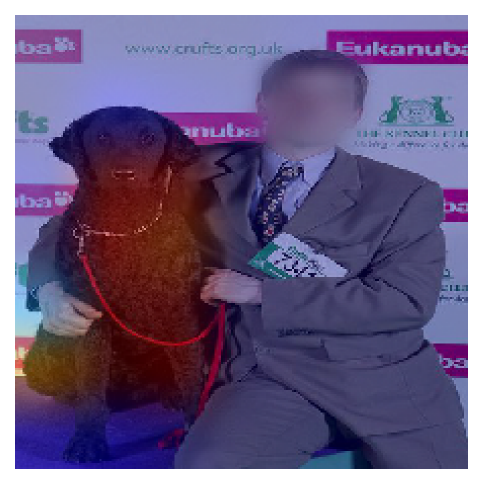}}
    \\
    \subfloat[Target class: ``Suit'']{\includegraphics[scale=0.37]{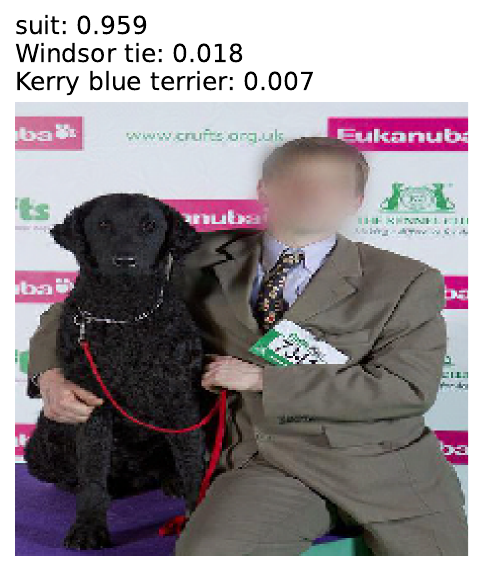}
    \hskip -7pt
    \includegraphics[scale=0.37]{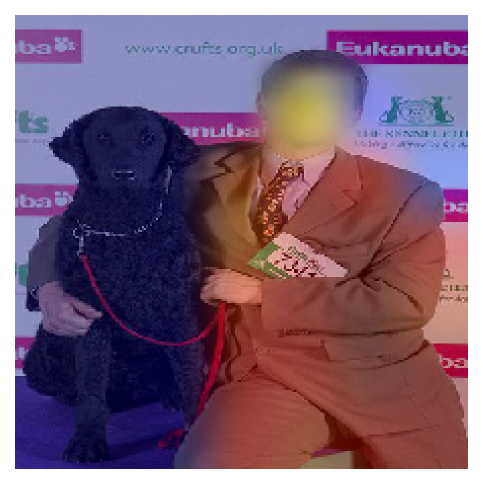}}
    \hskip 6pt
     \subfloat[Target class: ``Irish water spaniel'']{\includegraphics[scale=0.37]{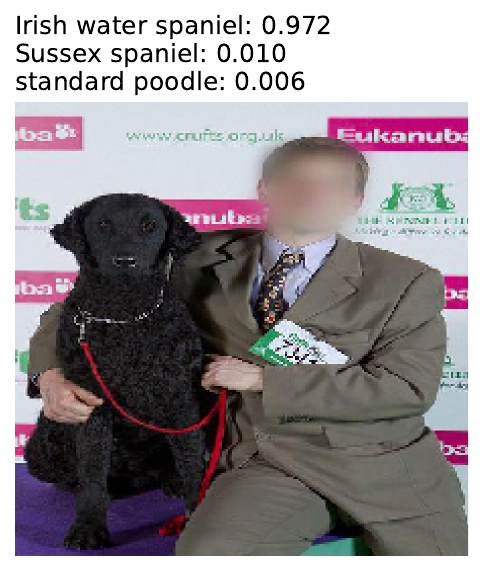}
    \hskip -7pt
    \includegraphics[scale=0.37]{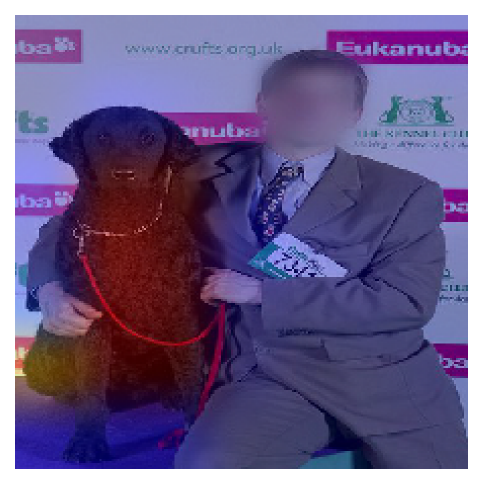}}
    \\
    \subfloat[``Curly-coated retriever'']{\includegraphics[scale=0.455]{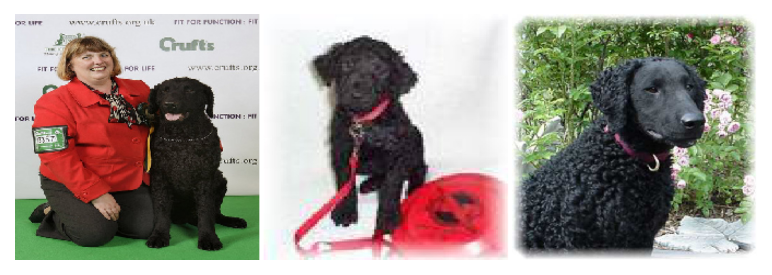}}
    \hskip 6pt
    \subfloat[``Irish water spaniel'']{\includegraphics[scale=0.455]{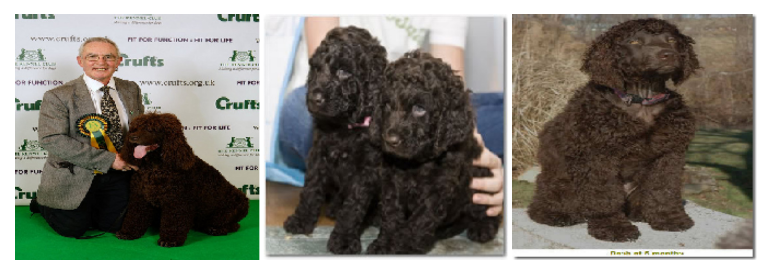}}
    \caption{Adversarial examples generated for the ImageNet dataset classification task, taking advantage of the class ambiguity introduced by the appearance of multiple concepts in the image. (a) Original input. 
    (b) {\mfrr Input perturbed in order to maximize confidence in the original class without altering the enhanced region in the explanation.}
    (c) Adversarial example targeting the class ``Suit'' and a target saliency-map highlighting the region in which this class appears.
(d) Adversarial example targeting the class ``Irish water spaniel'' and a target saliency-map highlighting the region in which the ground-truth class (``Curly-coated retriever'') appears. (e) \& (f) The 3 training images belonging to the class ``Curly-coated retriever'' and ``Irish water spaniel'', respectively, which are closest to the original input.}
    \label{fig:imagenet_dog_suit_adv}
\end{figure}

In Figure \ref{fig:imagenet_dog_suit_adv}, a different type of ambiguity will be considered to generate the adversarial examples: the appearance of multiple concepts or classes in the image. In such cases, adversarial examples can be employed to change the focus of the classification to one of the objects of interest. The explanation is, therefore, a key factor in order to further support the model decision in classifying the input as the class of interest selected by the adversary. 
The input in Figure \ref{fig:imagenet_dog_suit_adv}-(a) contains two classes that could be equally relevant: ``Curly-coated retriever'' dog breed (ground-truth class) and ``suit''. In Figures \ref{fig:imagenet_dog_suit_adv}-(b) and \ref{fig:imagenet_dog_suit_adv}-(c), adversarial examples are generated in order to ``untie'' this ambiguity, maximizing the confidence of one of the classes (``Curly-coated retriever'' and ``suit'', respectively) and changing the explanation to highlight the selected parts {\mfrr (paradigm A.2.1)}.\footnote{In this case, the target saliency maps have been generated using an image-segmentation model (Mask R-CNN with Inception Resnet v2), which has been used to segment the two desired parts. The pretrained model is accessible at \url{https://tfhub.dev/tensorflow/mask_rcnn/inception_resnet_v2_1024x1024/1}. {\mfrr Note that, in Figure \ref{fig:imagenet_dog_suit_adv}-(b), both the classification and the explanation are preserved, thus no attack is carried out.}}
As can be seen, the adversarial examples effectively focus the prediction on one of the classes, which can therefore bias the human interpretation of the result, accepting the prioritized output class as the dominant one.  Whereas this type of attacks are limited to the objects appearing in the image, different types of ambiguity can be considered at the same time to produce misclassifications that may be taken as ``correct'' for humans, as shown in Figure \ref{fig:imagenet_dog_suit_adv}-(d), in which the focus is not only placed on the dog, but also an incorrect class is produced (``Irish water spaniel'') taking advantage of the ambiguity of the class similarity {\mfrr (paradigm A.3)}. This ambiguity is, indeed, also reflected in the output confidence scores provided by the model when the original input is classified, shown above the left part of Figure \ref{fig:imagenet_dog_suit_adv}-(a), as both classes achieved a similar score. In order to assess the similarity between these two classes, figures \ref{fig:imagenet_dog_suit_adv}-(e) and \ref{fig:imagenet_dog_suit_adv}-(f) show the 3 prototypes belonging to each of the two classes that are the closest to the original image, in which it can be seen that both breeds contain very similar features.

\begin{figure}[!t]
    \centering
    \subfloat[]{
    \includegraphics[scale=0.365]{original_lab5-eps-converted-to.pdf} 
    \hskip -6pt
    \includegraphics[scale=0.365]{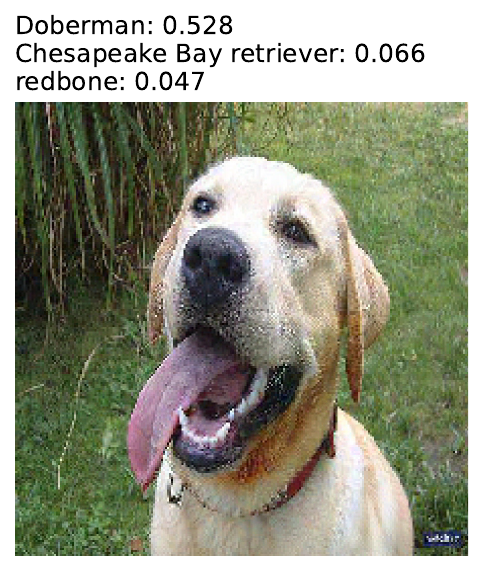}}
    \hskip 6pt
    \subfloat[]{
    \includegraphics[scale=0.44]{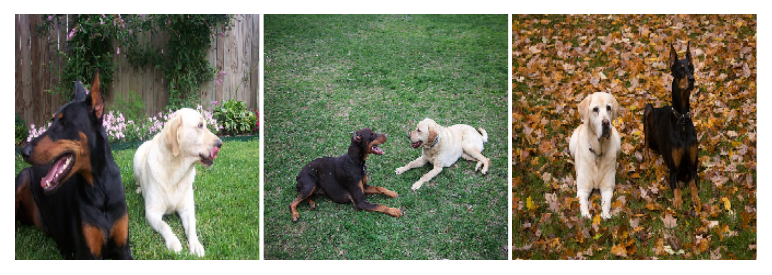}}
    \\
    \vspace{-0.3cm}
    \subfloat[]{
    \includegraphics[scale=0.44]{knn_clean_auto_lab5-eps-converted-to.pdf}}
    \hskip 6pt
    \subfloat[]{
    \includegraphics[scale=0.44]{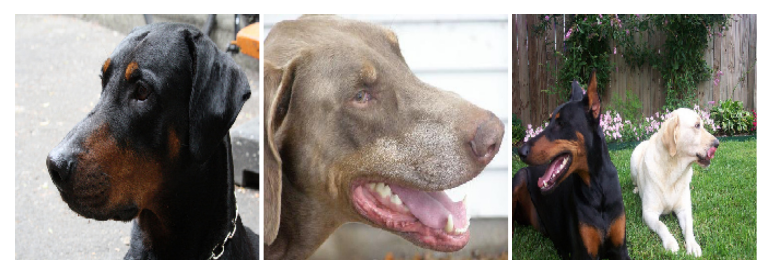}}
%\end{tabular}
    \caption{Adversarial example for the large scale visual recognition task, assuming a prototype-based explanation. (a) Original input belonging to the class ``Labrador Retriever'' (left) and adversarial example targetting the class ``Doberman'' (right). 
(b) Prototype-based explanation of the adversarial example and the class ``Doberman''  (that is, the 3 training images belonging to the class ``Doberman'' that are closest to the adversarial example).
(c) Prototype-based explanation of the original input and the ground-truth class.
(d) Prototype-based explanation of the original input and the target class ``Doberman''.
}
    \label{fig:adv_proto_exp}
\end{figure}

Finally, in Figure \ref{fig:adv_proto_exp}, we provide an illustrative example of an attack designed to fool a model whose decisions are explained using a prototype-based explanation. As discussed in Section \ref{sec:attacks_based_on_exp_types}, an adversary can take advantage of prototype-based explanations to support certain misclassifications, for instance, producing an incorrect output class and minimizing the distance with prototypes which, apart from containing features representative of the source class, are representative of the target class as well. Figure \ref{fig:adv_proto_exp}-(a) shows a well-classified image (left) and an adversarial example targeting the class ``Doberman''. The adversarial perturbation has been optimized in order to reduce the distance (in the latent representation) between the input and the 3 training images (belonging to the target class) shown in Figure \ref{fig:adv_proto_exp}-(b). As can be seen, these training images not only contain features representative of the target class (``Doberman''), but also additional features that resemble those in the original input sample (indeed, a similar dog is present in the selected training images), exemplifying the attack paradigm A.3.
Figure \ref{fig:adv_proto_exp}-(c) shows the prototypes belonging to the source class that are the closest to the original image, and \ref{fig:adv_proto_exp}-(d) those prototypes closest to the original image yet belonging to the target class. Note that both Figures \ref{fig:adv_proto_exp}-(b) and \ref{fig:adv_proto_exp}-(d) contain prototypes belonging to the target class, however, those which are adversarially produced appear considerably more coherent due to their ambiguity (in the sense that they contain prototypical features of both the source and target class).

\section{Conclusions}
\label{sec:conclusions}

In this paper, we have introduced a comprehensive framework to rigorously study the possibilities and limitations of adversarial examples in explainable machine learning scenarios, in which the input, the predictions of the models and the explanations are assessed by humans. First, we have extended the notion of adversarial examples in order to fit in such scenarios, which has allowed us to examine different adversarial attack paradigms. Furthermore,  we thoroughly analyze how adversarial attacks should be designed in order to mislead explainable machine learning models (and humans) depending on a wide range of factors such as the type of task addressed, the expertise of the users querying the model, as well as the type, scope or impact of the explanation methods used to justify the decisions of the models. Furthermore, the introduced attack paradigms have been illustrated using two representative image classification tasks and two different explanation methods based on feature-attribution explanations and example-based explanations.
Overall, the proposed framework provides a comprehensive road  map  for  the  design of  malicious  attacks  in realistic  scenarios involving explainable models and a human supervision, contributing to a more rigorous study of adversarial examples in the field of explainable machine learning.

\section{Future work}
\label{sec:future_work}

In this last section, we identify different promising research directions that could be derived from the contributions of our paper. First, additional factors could be considered in the proposed framework (jointly with the input, the output class and the explanation) in order to consider even more fine-grained scenarios, such as the confidence of the prediction, which can condition the human acceptance of a model prediction, as recently studied in \citet{nguyen2021effectiveness}. Moreover, an interesting research line could be developing a general and unifying attack algorithm capable of addressing all the attack paradigms described in our framework, that is, an approach capable of automatically generating adversarial examples which satisfy the most important requirements depending on the scenario, explanation method or attack paradigm that wants to be produced. We plan to study the generation of such attacks in future works. 

More generally, conceiving strategies to improve the reliability and robustness of explanation methods continues to be an urgent line of research, as still limited research has been conducted on the adversarial robustness of different explanation methods such as prototype-based approaches. Thus, a deeper analysis of the vulnerability of current explanation methods is an important step in order to increase the reliability and trustworthiness of explainable machine learning models.

\section*{Funding information}

This work was supported by the Basque Government (BERC 2022-2025 and ELKARTEK programs), by the Spanish Ministry of Economy and Competitiveness MINECO (project PID2019-104966GB-I00) and by the Spanish Ministry of Science, Innovation and Universities (FPU19/03231 predoctoral grant). Jose A. Lozano acknowledges support by the Spanish Ministry of Science, Innovation and Universities through BCAM Severo Ochoa accreditation (SEV-2017-0718).

\section*{Conflict of interest}
The authors have declared no conflicts of interest for this article.

\section*{Data availability statement}
The data that support the findings of this study are available at \url{https://github.com/vadel/AE4XAI}. These data were derived from the following resources available at \url{https://www.image-net.org/} and \url{https://github.com/lindawangg/COVID-Net}.

\bibliographystyle{apalike}
\bibliography{mybibfile}
%\printbibliography

\end{document}